\documentclass[10pt,twocolumn,letterpaper]{article}

\usepackage{cvpr}              %

\usepackage{pifont}%
\newcommand{\cmark}{\ding{51}}%
\newcommand{\xmark}{\ding{55}}%
\usepackage{tabularx}

\usepackage{appendix}
\usepackage{graphicx}
\usepackage{booktabs}
\usepackage{wrapfig}
\usepackage[utf8]{inputenc} %
\usepackage[T1]{fontenc}    %
\usepackage{url}            %
\usepackage{booktabs}       %
\usepackage{amsfonts}       %
\usepackage{nicefrac}       %
\usepackage{microtype}      %
\usepackage{xcolor}         %

\usepackage[accsupp]{axessibility}
\usepackage{multirow}
\usepackage{tabu}
\usepackage{tabularray}
\usepackage{colortbl}

\definecolor{tabfirst}{rgb}{1, 0.7, 0.7}
\definecolor{tabsecond}{rgb}{1, 0.85, 0.7}
\definecolor{tabthird}{rgb}{1, 1, 0.7}

\definecolor{purple}{RGB}{138, 43, 226}

\definecolor{cvprblue}{rgb}{0.21,0.49,0.74}

\usepackage{colortbl}
\definecolor{tabfirst}{rgb}{1, 0.7, 0.7}
\definecolor{tabsecond}{rgb}{1, 0.85, 0.7}
\definecolor{tabthird}{rgb}{1, 1, 0.7}
\usepackage[pagebackref,breaklinks,colorlinks,allcolors=cvprblue]{hyperref}

\def\paperID{9297} %
\def\confName{CVPR}
\def\confYear{2025}

\title{A Lightweight UDF Learning Framework for 3D Reconstruction Based on Local Shape Functions}

\author{Jiangbei Hu$^{1,2*}$\quad Yanggeng Li$^{2}$\thanks{Equal contribution}\quad Fei Hou$^{3,4}$\quad Junhui Hou$^{5}$\quad Zhebin Zhang$^{6}$\quad Shengfa Wang$^{1}$\\  Na Lei$^{1}$\quad Ying He$^{2}$\thanks{Corresponding author: yhe@ntu.edu.sg}\\
$^{1}$ School of Software, Dalian University of Technology\quad
$^{2}$ CCDS, Nanyang Technological University\\
$^{3}$ Key Laboratory of System Software (CAS) and SKLCS, Institute of Software, CAS\\
$^{4}$ University of Chinese Academy of Sciences \\
$^{5}$ Department of Computer Science, City University of Hong Kong\quad
$^{6}$ OPPO\\
}

\begin{document}
\maketitle
\begin{abstract}
    Unsigned distance fields (UDFs) provide a versatile framework for representing a diverse array of 3D shapes, encompassing both watertight and non-watertight geometries. Traditional UDF learning methods typically require extensive training on large 3D shape datasets, which is costly and necessitates re-training for new datasets. This paper presents a novel neural framework, \textit{LoSF-UDF}, for reconstructing surfaces from 3D point clouds by leveraging local shape functions to learn UDFs. We observe that 3D shapes manifest simple patterns in localized regions, prompting us to develop a training dataset of point cloud patches characterized by mathematical functions that represent a continuum from smooth surfaces to sharp edges and corners. Our approach learns features within a specific radius around each query point and utilizes an attention mechanism to focus on the crucial features for UDF estimation. Despite being highly lightweight, with only 653 KB of trainable parameters and a modest-sized training dataset with 0.5 GB storage, our method enables efficient and robust surface reconstruction from point clouds without requiring for shape-specific training. Furthermore, our method exhibits enhanced resilience to noise and outliers in point clouds compared to existing methods. We conduct comprehensive experiments and comparisons across various datasets, including synthetic and real-scanned point clouds, to validate our method's efficacy. Notably, our lightweight framework offers rapid and reliable initialization for other unsupervised iterative approaches, improving both the efficiency and accuracy of their reconstructions. Our project and code are available at  
\href{https://jbhu67.github.io/LoSF-UDF.github.io/}{https://jbhu67.github.io/LoSF-UDF.github.io/}.

\end{abstract}
    
\section{Introduction}
\label{sec:intro}

3D surface reconstruction from raw point clouds is a significant and long-standing problem in computer graphics and machine vision. Traditional techniques like Poisson Surface Reconstruction~\cite{2006Poisson} create an implicit indicator function from oriented points and reconstruct the surface by extracting an appropriate isosurface. The advancement of artificial intelligence has led to the emergence of numerous neural network-based methods for 3D reconstruction. Among these, neural implicit representations have gained significant influence, which utilize signed distance fields (SDFs)~\cite{Park_2019_CVPR,chabra2020deep,NeuralPull,Wang_2022,Liu2021MLS,neuralimls2023wang, On-SurfacePriors} and occupancy fields ~\cite{Occupancy_Networks,chibane20ifnet,Peng2020ECCV,boulch2022poco} to implicitly depict 3D geometries. SDFs and occupancy fields extract isosurfaces by solving regression and classification problems, respectively. However, both techniques require internal and external definitions of the surfaces, limiting their capability to reconstructing only watertight geometries. Therefore, unsigned distance fields~\cite{chibane2020ndf,zhou2023levelset,ren2023geoudf,ye2022gifs,Zhou2022CAP-UDF,li2023neural, fainstein2024dudf, UODFs} have recently gained increasing attention due to their ability to reconstruct non-watertight surfaces and complex geometries with arbitrary topologies.

Reconstructing 3D geometries from raw point clouds using UDFs presents significant challenges due to the non-differentiability near the surface. This characteristic complicates the development of loss functions and undermines the stability of neural network training. Various unsupervised approaches~\cite{Zhou2022CAP-UDF, zhou2023levelset, fainstein2024dudf} have been developed to tailor loss functions that leverage the intrinsic characteristics of UDFs, ensuring that the reconstructed geometry aligns closely with the original point clouds. However, these methods suffer from slow convergence, necessitating an extensive network training time to reconstruct a single geometry. As a supervised method, GeoUDF~\cite{ren2023geoudf} learns local geometric priors through training on datasets such as ShapeNet~\cite{shapenet2015}, thus achieving efficient UDF estimation. Nonetheless, the generalizability of this approach is dependent on the training dataset, which also leads to relatively high computational costs.

In this paper, we propose a lightweight and effective supervised learning framework, \textit{LoSF-UDF}, to address these challenges. Since learning UDFs does not require determining whether a query point is inside or outside the geometry, it is a local quantity independent of the global context. Inspired by the observation that 3D shapes manifest simple patterns within localized areas, we synthesize a training dataset comprising a set of point cloud patches by utilizing local shape functions. 
Subsequently, we can estimate the unsigned distance values by learning local geometric features through an attention-based network.
Our approach distinguishes itself from existing methods by its novel training strategy. Specifically, it is uniquely trained on synthetic surfaces, yet it demonstrates remarkable capability in predicting UDFs for a wide range of common surface types. For smooth surfaces, we generate training patches (quadratic surfaces) by analyzing principal curvatures, meanwhile, we design simple shape functions to imitate sharp features. This strategy has three unique advantages. First, it systematically captures the local geometries of most common surfaces encountered during testing, effectively mitigating the dataset dependence risk that plagues current UDF learning methods. Second, for each training patch, the ground-truth UDF is readily available, streamlining the training process. Third, this approach substantially reduces the costs associated with preparing the training datasets. We evaluate our framework on various datasets and demonstrates its ability to robustly reconstruct high-quality surfaces, even for point clouds with noise and outliers. Notably, our method can serve as a lightweight initialization that can be integrated with existing unsupervised methods to enhance their performance.
We summarize our main contributions as follows.
\begin{itemize}
\item We present a simple yet effective data-driven approach that learns UDFs directly from a synthetic dataset consisting of point cloud patches, which is independent of the global shape.
    
\item Our method is computationally efficient and requires training only once on our synthetic dataset. Then it can be applied to reconstruct a wide range of surface types.

\item Our lightweight framework offers rapid and reliable initialization for other unsupervised iterative approaches, improving both efficiency and accuracy.

\end{itemize}

\begin{figure*}[!t]
    \centering
    \includegraphics[width=1.0\linewidth]{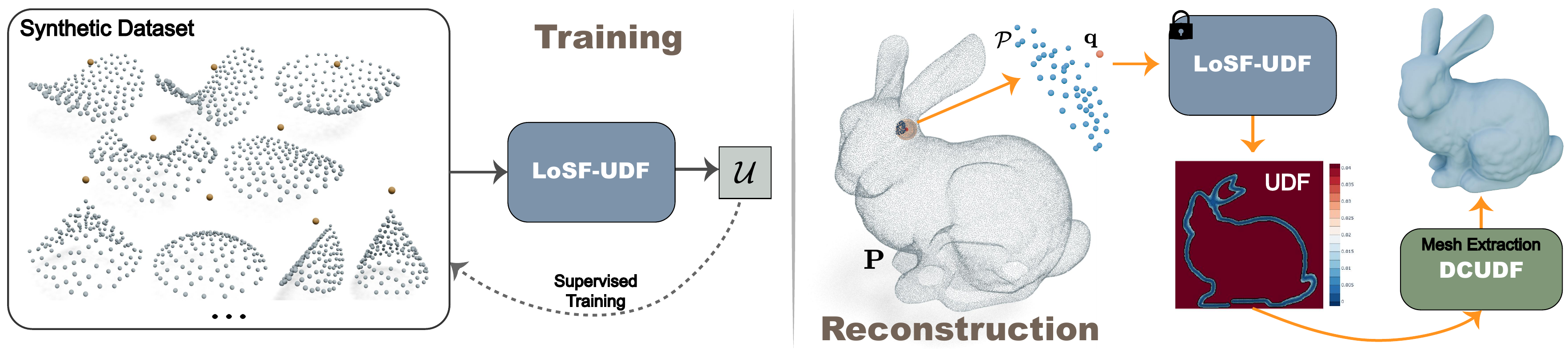}
    \vspace{-0.2in}
    \caption{Pipeline. First, we train a UDF prediction network $\mathcal{U}_\Theta$ on a synthetic dataset, which contains a series of local point cloud patches that are independent of specific shapes.  Given a global point cloud $\mathbf{P}$, we then extract a local patch $\mathcal{P}$ assigned to each query point $\mathbf{q}$ within a specified radius, and obtain the corresponding UDF values $\mathcal{U}_{\hat{\Theta}}(\mathcal{P}, \mathbf{q})$. Finally, we extract the mesh corresponding to the input point cloud by incorporating the DCUDF~\cite{Hou2023DCUDF} framework. }
    \vspace{-0.22in}
    \label{fig:pipeline}
\end{figure*}
\section{Related work}
\vspace{-0.1in}

\textbf{Neural surface representations.} Reconstructing 3D surfaces from point clouds is a classic and important topic in computer graphics~\cite{berger2014state,berger2017survey,xu2024parameterization}. Recently, the domain of deep learning has spurred significant advances in the implicit neural representation of 3D shapes. Some of these works trained a classifier neural network to construct occupancy fields~\cite{Occupancy_Networks,chibane20ifnet,Peng2020ECCV,boulch2022poco} for representing 3D geometries.
Poco~\cite{boulch2022poco} achieves superior reconstruction performance by introducing convolution into occupancy fields. Ouasfi \etal~\cite{ouasfi2024unsupervised} recently proposed a uncertainty measure method based on margin to learn occupancy fields from sparse point clouds.
Compared to occupancy fields, SDFs~\cite{Park_2019_CVPR,NeuralPull,Wang_2022,Liu2021MLS,neuralimls2023wang, On-SurfacePriors} offer a more precise geometric representation by differentiating between interior and exterior spaces through the assignment of signs to distances. 

\noindent\textbf{Unsigned distance fields learning.}
Although Occupancy fields and SDFs have undergone significant development recently, they are hard to reconstruct surfaces with boundaries or nonmanifold features. G-Shell~\cite{liu2024gshell} developed a differentiable shell-based representation for both watertight and non-watertight surfaces. However, UDFs provide a simpler and more natural way to represent general shapes~\cite{chibane2020ndf,zhou2023levelset,ren2023geoudf,ye2022gifs,Zhou2022CAP-UDF,li2023neural, fainstein2024dudf, UODFs}. Various methods have been proposed to reconstruct surfaces from point clouds by learning UDFs. CAP-UDF~\cite{Zhou2022CAP-UDF} suggested directing 3D query points toward the surface with a consistency constraint to develop UDFs that are aware of consistency. LevelSetUDF~\cite{zhou2023levelset} learned a smooth zero-level function within UDFs through level set projections. As a supervised approach, GeoUDF~\cite{ren2023geoudf} estimates UDFs by learning local geometric priors from training on many 3D shapes.
DUDF~\cite{fainstein2024dudf} formulated the UDF learning as an Eikonal problem with distinct boundary conditions. UODF~\cite{UODFs} proposed unsigned orthogonal distance fields that every point in this field can access the closest surface points along three orthogonal directions. Instead of reconstructing from point clouds, many recent works~\cite{deng20242sudf, meng_2023_neat, long2023neuraludf, Liu23NeUDF} learn high-quality UDFs from multi-view images to reconstruct non-watertight surfaces. Furthermore, UiDFF~\cite{udiff} presents a 3D diffusion model for UDFs to generate textured 3D shapes with boundaries. 

\noindent\textbf{Local-based reconstruction.} Most methods achieve 3D reconstruction by constructing a global function from point clouds. For example, Poisson methods~\cite{2006Poisson,2013Screened} fits surfaces by solving partial differential equations, while neural network-based methods like DeepSDF~\cite{chen2019learning,mescheder2019occupancy,park2019deepsdf} represent geometry through network optimization. The limitation of most global methods lies in their need for extensive datasets for training, coupled with inadequate generalization to unseen shape categories. Conversely, 3D surfaces exhibit local similarities and repetitions, which have spurred the development of techniques for reconstructing surfaces locally. Ohtake \etal ~\cite{ohtake2005multi} introduced a shape representation utilizing a multi-scale partition of unity framework, wherein the local shapes of surfaces are characterized by piecewise quadratic functions. DeepLS~\cite{chabra2020deep} and LDIF~\cite{genova2020local} reconstructed local SDF by training learnable implicit functions or neural networks. PatchNets~\cite{tretschk2020patchnets} proposed a mid-level patch-based surface representation, facilitating the development of models with enhanced generalizability. Ying \etal ~\cite{ying2023adaptive} introduced a local-to-local shape completion framework that utilized adaptive local basis functions. While these methods all focus on SDF, GeoUDF~\cite{ren2023geoudf} represents a recent advancement in reconstructing UDF from a local perspective. 

\section{Method}
\textbf{\textbf{Motivation.}} Distinct from SDFs, there is no need for UDFs to determine the sign to distinguish between the inside and outside of a shape. Consequently, the UDF values are solely related to the local geometric characteristics of 3D shapes. Furthermore, within a certain radius for a query point, local geometry can be approximated by general mathematical functions. Stemming from these insights, we propose a novel UDF learning framework that focuses on local geometries. We employ local shape functions to construct a series of point cloud patches as our training dataset, which includes common smooth and sharp geometric features. Given a point cloud to reconstruct, we employ the optimized model to output the corresponding distance values based on the local patch within radius for each query point.
\Cref{fig:pipeline} illustrates the pipeline of our proposed UDF learning framework.

\subsection{Local shape functions}
\begin{figure}[!b]
    \centering
    \includegraphics[width=0.48\linewidth]{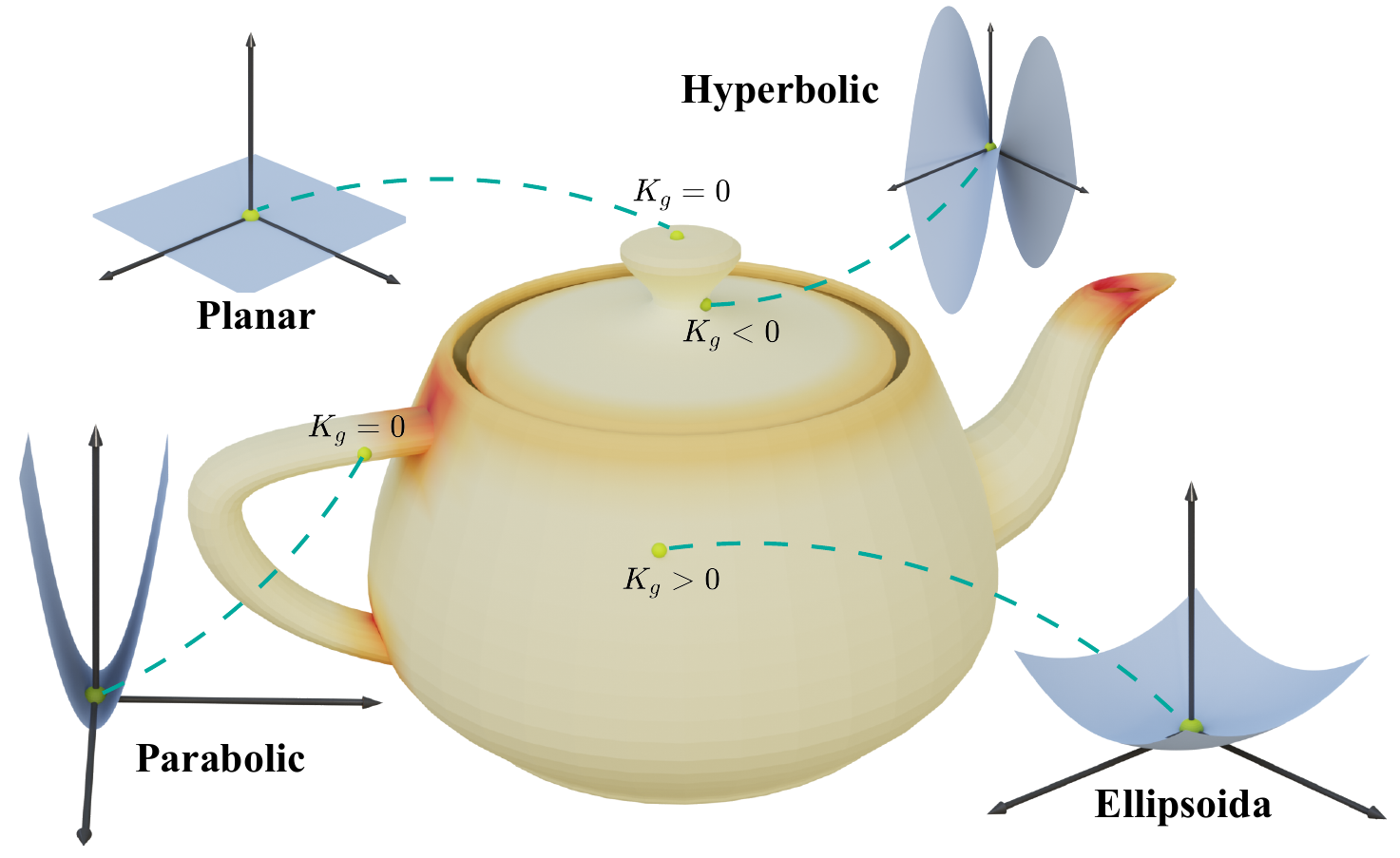}
    \hfill
    \includegraphics[width=0.48\linewidth]{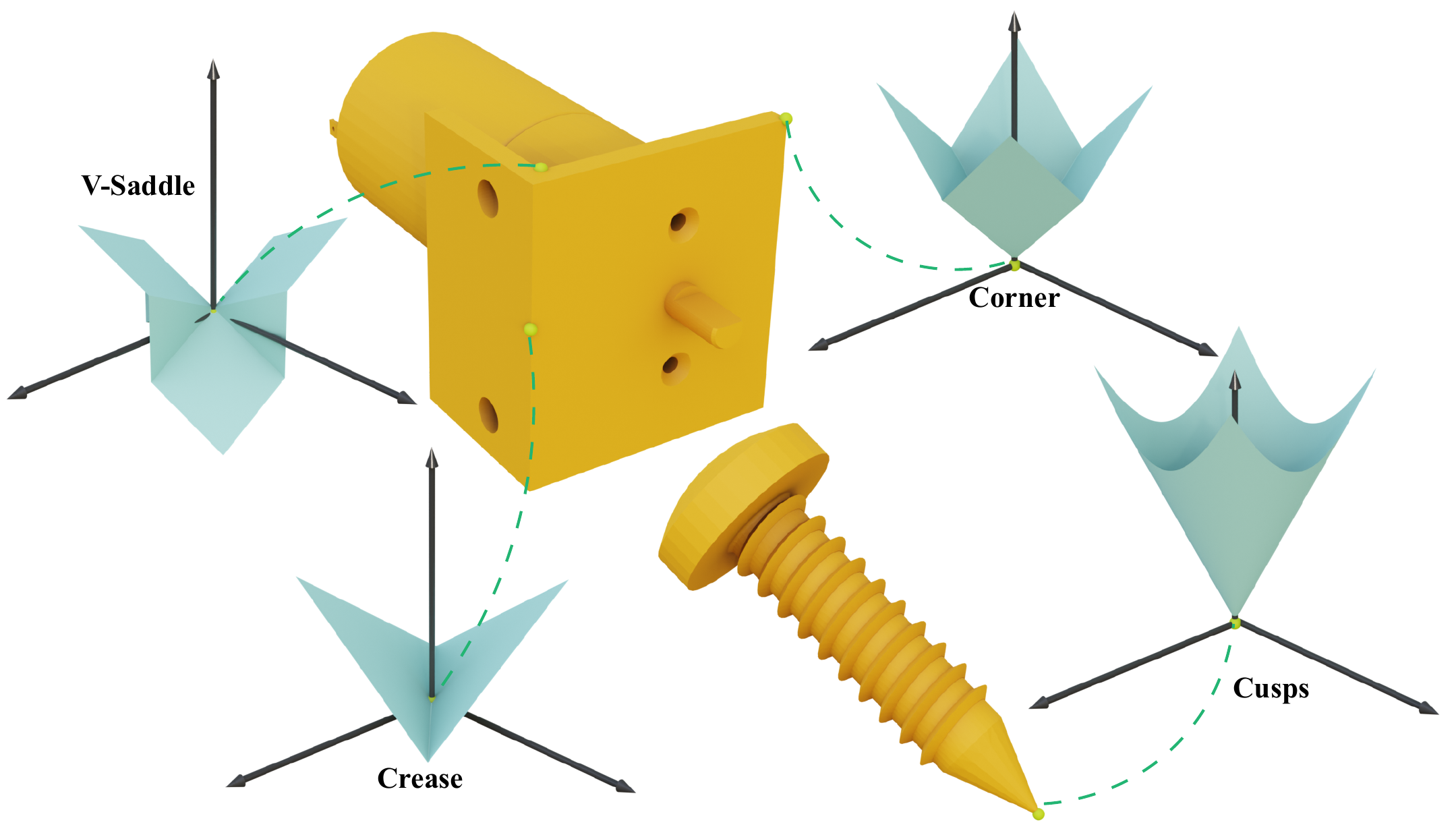}
    \\
    \makebox[0.48\linewidth]{(a) Smooth patches}
    \hfill
    \makebox[0.48\linewidth]{(b) Sharp patches}
    \vspace{-0.1in}
    \caption{Local geometries. (a) For points on a geometry that are differentiable, the local shape at these points can be approximated by quadratic surfaces. (b) For points that are non-differentiable, we can also construct locally approximated surfaces using functions.} 
    \label{fig:local-geo}
    \vspace{-0.2in}
\end{figure}
\textbf{Smooth patches.}
From the viewpoint of differential geometry~\cite{do2016differential}, the local geometry at a specific point on a regular surface can be approximated by a quadratic surface. Specifically, consider a regular surface $\mathcal{S}: \mathbf{r}=\mathbf{r}(u, v)$ with a point $\mathbf{p}$ on it. At point $\mathbf{p}$, it is possible to identify two principal direction unit vectors, $\mathbf{e}_1$ and $\mathbf{e}_2$, with the corresponding normal $\mathbf{n}=\mathbf{e}_1\times\mathbf{e}_2$. A suitable parameter system $(u,v)$ can be determined such that $\mathbf{r}_u=\mathbf{e}_1$ and $\mathbf{r}_v=\mathbf{e}_2$, thus obtaining the corresponding first and second fundamental forms as\vspace{-0.1in}
\begin{equation}\label{eq:fundmental-12}
\setlength{\abovedisplayskip}{7pt}
\setlength{\belowdisplayskip}{7pt}
    [\mathrm{I}]_{\mathbf{p}}=\begin{bmatrix}
        1 &0\\
        0 &1
    \end{bmatrix},\quad
    [\mathrm{II}]_{\mathbf{p}}=\begin{bmatrix}
        \kappa_1 &0\\
        0 &\kappa_2
    \end{bmatrix},
\end{equation}
where $\kappa_1, \kappa_2$ are principal curvatures. 
Without loss of generality, 
we assume $\mathbf{p}$ corresponding to $u=v=0$ and expand the Taylor form at this point as
\begin{equation}
\setlength{\abovedisplayskip}{4pt}
\setlength{\belowdisplayskip}{4pt}
\label{eq:taylor}
\begin{aligned}
     &\mathbf{r}(u,v)=\mathbf{r}(0,0)+\mathbf{r}_u(0,0)u+\mathbf{r}_v(0,0)v+\frac{1}{2}[\mathbf{r}_{uu}(0,0)u^2\\ &+
     \mathbf{r}_{uv}(0,0)uv+\mathbf{r}_{vv}(0,0)v^2]+o(u^2+v^2).
\end{aligned}
\end{equation}
Decomposing $\mathbf{r}_{uu}(0,0), \mathbf{r}_{uv}(0,0), \text{and } \mathbf{r}_{vv}(0,0)$ along the tangential and normal directions, we can formulate \cref{eq:taylor} according to \cref{eq:fundmental-12} as
\begin{equation}
\setlength{\abovedisplayskip}{4pt}
\setlength{\belowdisplayskip}{4pt}
    \begin{aligned}
        &\mathbf{r}(u,v)=\mathbf{r}(0,0)+(u+o(\sqrt{u^2+v^2}))\mathbf{e}_1+(v\\
        &+o(\sqrt{u^2+v^2}))\mathbf{e}_2+\frac{1}{2}(\kappa_1u^2+\kappa_2v^2+o(u^2+v^2)))\mathbf{n}
    \end{aligned}
\end{equation}
where $o(u^2+v^2)\approx 0$ is negligible in a small local region. Consequently, by adopting $\{\mathbf{p}, \mathbf{e}_1, \mathbf{e}_2, \mathbf{n}\}$ as the orthogonal coordinate system, we can define the form of the local approximating surface as
\begin{equation}
\setlength{\abovedisplayskip}{4pt}
\setlength{\belowdisplayskip}{4pt}
\label{eq:smooth}
    x=u,\quad y=v,\quad z=\frac{1}{2}(\kappa_1u^2+\kappa_2v^2),
\end{equation}
which exactly are quadratic surfaces $z=\frac{1}{2}(\kappa_1x^2+\kappa_2y^2)$. Furthermore, in relation to Gaussian curvatures $\kappa_1\kappa_2$, quadratic surfaces can be categorized into four types: ellipsoidal, hyperbolic, parabolic, and planar. As shown in \cref{fig:local-geo}, for differentiable points on a general geometry, the local shape features can always be described by one of these four types of quadratic surfaces.

\begin{figure}[!t]
    \centering
    \includegraphics[width=1.0\linewidth]{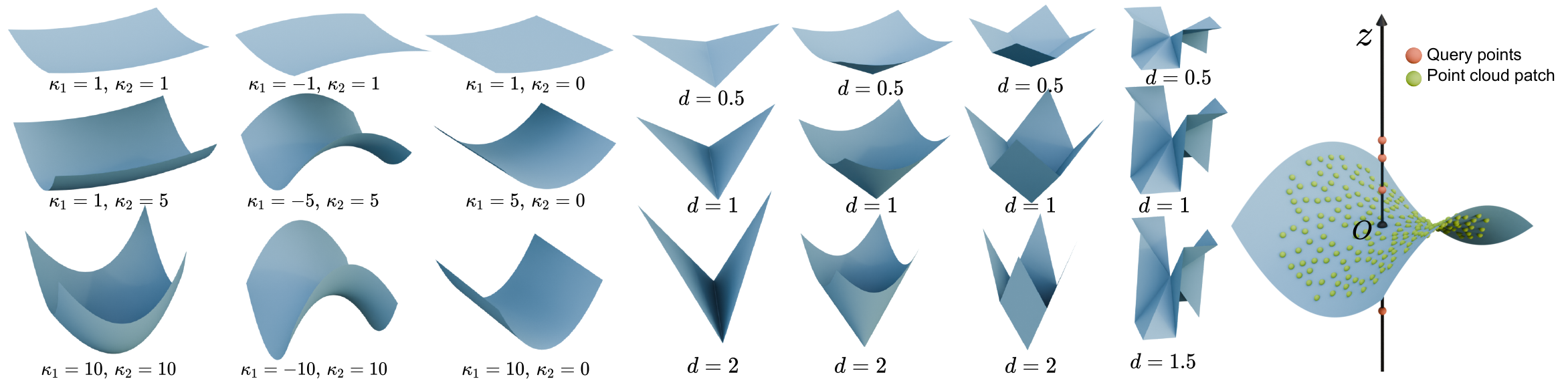}
    \vspace{-0.2in}
    \caption{Synthetic surfaces for training. By manipulating functional parameters, we can readily create various smooth and sharp surfaces, subsequently acquiring pairs of point cloud patches and query points via sampling.}
    \vspace{-0.2in}
    \label{fig:syn-data}
\end{figure}

\noindent\textbf{Sharp patches.}
For surfaces with sharp features, they are not differentiable at some points and cannot be approximated in the form of a quadratic surface.
We categorize commonly seen sharp geometric features into four types, including creases, cusps, corners, and v-saddles, as illustrated in \cref{fig:local-geo}(b). We construct these four types of sharp features in a consistent form $z=f(x,y)$ like smooth patches. We define a family of functions as 
\begin{equation}
\setlength{\abovedisplayskip}{5pt}
\setlength{\belowdisplayskip}{5pt}
\label{eq:sharp}
    z=1-h\cdot g(x,y), 
\end{equation}
where $h$ can adjust the sharpness of the shape. Specifically, $g=\frac{|kx-y|}{\sqrt{1+k^2}}$ for creases ($k$ can control the direction), $g=\sqrt{x^2+y^2}$ for cusps, $g=\max (|x|, |y|)$ for corners, $g=(|x|+|y|)\cdot(\frac{|x|}{x}\cdot\frac{|y|}{y})$ for v-saddles.
\cref{fig:syn-data} illustrates several examples of smooth and sharp patches with distinct parameters.

\noindent\textbf{Synthetic training dataset.}
We utilize the mathematical functions introduced above to synthesize a series of point cloud patches for training. As shown in \cref{fig:syn-data}, we first uniformly sample $m$ points $\{(x_i, y_i)\}_{i=1}^m$ within a circle of radius $r_0$ centered at $(0, 0)$ in the $xy$-plane. Then, we substitute the coordinates into \cref{eq:smooth,eq:sharp} to obtain the corresponding $z$-coordinate values, resulting in a patch $\mathcal{P}=\{\mathbf{p}_{i=1}^m\}$, where $\mathbf{p}_i=(x_i, y_i, z(x_i, y_i))$. Subsequently, we randomly collect query points $\{\mathbf{q_i}\}_{i=1}^n$ distributed along the vertical ray intersecting the $xy$-plane at the origin, extending up to a distance of $r_0$. For each query point $\mathbf{q}_i$, we determine its UDF value $\mathcal{U}(\mathbf{q}_i)$, which is either $|\mathbf{q}_i^{(z)}|$ for smooth patches or $1-|\mathbf{q}_i^{(z)}|$ for sharp patches. Noting that for patches with excessively high curvature or sharpness, the minimum distance of the query points may not be the distance to $(0, 0, z(0, 0))$, we will exclude these patches from our training dataset. Overall, each sample in our synthetic dataset is specifically in the form of $\{\mathbf{q}, \mathcal{P}, \mathcal{U}(\mathbf{q})\}$.

\subsection{UDF learning}
We perform supervised training on the synthesized dataset which is independent of specific shapes. The network learns the features of local geometries and utilizes an attention-based module to output the corresponding UDF values from the learned features. After training, given any 3D point clouds and a query point in space, we extract the local point cloud patch near the query, which has the same form as the data in the training dataset. Consequently, our network can predict the UDF value at that query point based on this local point cloud patch.

\subsubsection{Network architecture}
\begin{figure}[!b]
    \centering
    \vspace{-0.12in}
    \includegraphics[width=1\linewidth]{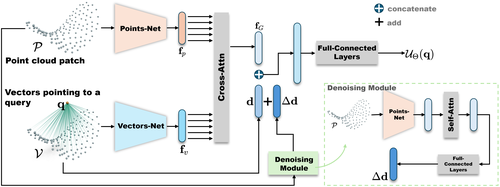}
    \vspace{-0.2in}
    \caption{Network architecture of LoSF-UDF.}
    \label{fig:network}
    \vspace{-0.1in}
\end{figure}

For a sample $\{\mathbf{q}, \mathcal{P}=\{\mathbf{p}_i\}_{i=1}^m, \mathcal{U}(\mathbf{q})\}$, we first obtain a latent code $\mathbf{f}_p\in\mathbb{R}^{l_p}$ related to the local point cloud patch $\mathcal{P}$ through a Point-Net~\cite{qi2017} $\mathcal{F}_p$. To derive features related to distance, we use relative vectors from the patch points to the query point, $\mathcal{V}=\{\mathbf{p}_i-\mathbf{q}\}_{i=1}^m$, as input to a Vectors-Net $\mathcal{F}_v$, which is similar to the Point-Net $\mathcal{F}_p$. This process results in an additional latent code $\mathbf{f}_v\in\mathbb{R}^{l_v}$. Subsequently, we apply a cross-attention module~\cite{vaswani2023attention} to obtain the feature codes for the local geometry,
\begin{equation}
    \mathbf{f}_G=\text{CrossAttn}(\mathbf{f}_p, \mathbf{f}_v)\in\mathbb{R}^{l_G},
\end{equation}
where we take $\mathbf{f}_p$ as the Key-Value (KV) pair and $\mathbf{f}_v$ as the Query (Q). In our experiments, we set $l_p=l_v=64$, and $l_G=128$. Based on the learned geometric features, we aim to fit the UDF values from the distance within the local point cloud. Therefore, we concatenate the distances $\mathbf{d}\in\mathbb{R}^m$ induced from $\mathcal{V}$ with the latent code $\mathbf{f}_G$, followed by a series of fully connected layers to output the predicted UDF values $\mathcal{U}_\Theta(\mathbf{q})$. \Cref{fig:network} illustrates the overall network architecture and data flow. The two PointNets used in our network to extract features from point cloud patches $\mathcal{P}$ and vectors $\mathcal{V}$ consist of four ResNet blocks. In addition, the two fully connected layer modules in our framework consist of three layers each.  To ensure non-negativity of the UDF values output by the network, we employ the softplus activation function.

\noindent\textbf{Denoising module.} 
In our network, even if point cloud patches are subjected to a certain degree of noise or outliers, their representations in the feature space should remain similar.
However, distances induced directly from noisy vectors $\mathcal{V}$ will inevitably contain errors, which can affect the accurate prediction of UDF values. To mitigate this impact, we introduce a denoising module  that predicts displacements $\Delta\mathbf{d}$ from local point cloud patches, as shown in \cref{fig:network}. We then add the displacements $\Delta\mathbf{d}$ to the distances $\mathbf{d}$ to improve the accuracy of the UDF estimation.

\subsubsection{Training and evaluation}
\textbf{Data augmentation.} 
During the training process, we scale all pairs of local patches $\mathcal{P}$ and query points $\mathbf{q}$ to conform to the bounding box constraints of $[-0.5, 0.5]$, and the corresponding GT UDF values $\mathcal{U}(\mathbf{q})$ are scaled by equivalent magnitudes. Given the uncertain orientation of local patches extracted from a specified global point cloud, we have applied data augmentation via random rotations to the training dataset. Furthermore, to enhance generalization to open surfaces with boundaries, we randomly truncate $20\%$ of the smooth patches to simulate boundary cases. To address the issue of noise handling, we introduce Gaussian noise $\mathcal{N}(0, 0.1)$ to $30\%$ of the data in each batch during every training epoch.

\noindent\textbf{Loss functions.} 
We employ $L_1$ loss $\mathcal{L}_{\text{u}}$ to measure the discrepancy between the predicted UDF values and the GT UDF values. Moreover, for the displacements $\Delta\mathbf{d}$ output by the denoising module, we employ $L_1$ regularization to encourage sparsity. Consequently, we train the network driven by the loss function $\mathcal{L}=\mathcal{L}_u+\lambda_d\mathcal{L}_r$,
where $\mathcal{L}_u=|\mathcal{U}(\mathbf{q})-\mathcal{U}_\Theta(\mathbf{q})|,\,\,\mathcal{L}_r=|\Delta\mathbf{d}|$, we set $\lambda_d=0.01$ in our experiments.

\noindent\textbf{Evaluation.} 
Given a 3D point cloud $\mathbf{P}$ for reconstruction, we first normalize it to fit within a bounding box with dimensions ranging from $[-0.5, 0.5]$. Subsequently, within the bounding box space, we uniformly sample grid points at a specified resolution to serve as query points. Finally, we extract the local geometry $\mathcal{P}_\mathbf{p}$ for each query point by collecting points from the point cloud that lie within a sphere of a specified radius centered on the query point. We can obtain the predicted UDF values by the trained network $\mathcal{U}_{\Theta^*}(\mathbf{q}, \mathcal{P}_\mathbf{q})$, where $\Theta^*$ represents the optimized network parameters. Note that for patches $\mathcal{P}_\mathbf{p}$ with fewer than 5 points, we set the UDF values as a large constant. Finally, we extract meshes from the UDFs using the DCUDF model~\cite{Hou2023DCUDF}.

\subsection{Integration with unsupervised methods}
Unsupervised methods, such as CAP-UDF~\cite{Zhou2022CAP-UDF} and LevelSetUDF~\cite{zhou2023levelset}, require time-consuming iterative reconstruction of a single point cloud. In contrast, our LoSF-UDF method is a highly lightweight framework. Once trained on a synthetic, shape-independent local patch dataset, it efficiently reconstructs plausible 3D shapes from diverse point clouds, even in the presence of noise and outliers. Although unsupervised methods are time-consuming, they can reconstruct shapes with richer details due to the combined effects of various loss functions. Therefore, we integrate our method with unsupervised approaches to provide better initialization, thereby accelerating convergence (\cref{tab:time}) and achieving improved reconstruction results.
Generally, assuming the network of the unsupervised method is $\mathcal{B}_\Xi$, we define the loss function of our integrated framework as 
\begin{equation}
    \label{eq:integration-loss}
    \min\limits_\Xi\,\mathcal{L}=\alpha_t\frac{1}{N}\sum_{i=1}^N|\mathcal{B}_\Xi(\mathbf{q}_i)-\mathcal{U}_{\Theta^*}(\mathbf{q}_i)|+(1-\alpha_t)\mathcal{L}_{\text{unsupv}},
\end{equation}
where $\mathcal{B}_\Xi$ can be selected as a MLP network like CAP-UDF~\cite{Zhou2022CAP-UDF} and LevelSetUDF~\cite{zhou2023levelset}, or a SIREN network~\cite{siren} like DEUDF~\cite{xu2024detail}. $\mathcal{L}_{\text{unsupv}}$ is the loss functions employed in these unsupervised methods. $\mathcal{U}_{\Theta^*}$ is our trained LoSF network with optimized parameters $\Theta^*$. $\alpha_t\in[0, 1]$ is a time-dependent weight. In our experiments (refer to \cref{sec:integration}), the whole training process requires around 20000 iterations. The value of $\alpha_t$ decreases from 1 to 0 gradually during the first 10000 iterations.

\section{Experimental results}

\subsection{Setup}

\textbf{Datasets.}
To compare our method with other state-of-the-art UDF learning approaches, we tested it on various datasets that include general artificial objects from the field of computer graphic. Following previous works~\cite{Liu23NeUDF, Zhou2022CAP-UDF, zhou2023levelset}, we select the "Car" category from ShapeNet~\cite{shapenet2015}, which has a rich collection of multi-layered and non-closed shapes. Furthermore, we select the real-world dataset DeepFashion3D~\cite{liuLQWTcvpr16DeepFashion} for open surfaces, and ScanNet~\cite{dai2017scannet} for large outdoor scenes. To assess our model's performance on actual noisy inputs, we conducted tests on real range scan dataset~\cite{10.1145/2451236.2451246} following the previous works~\cite{Zhou2022CAP-UDF, zhou2023levelset}.

\noindent\textbf{Baselines \& metrics.}
For our validation datasets, we compared our method against the state-of-the-art UDF learning models, which include unsupervised methods like CAP-UDF~\cite{Zhou2022CAP-UDF}, LevelSetUDF~\cite{zhou2023levelset}, and DUDF~\cite{fainstein2024dudf}, as well as the supervised learning method, GeoUDF~\cite{ren2023geoudf}. We trained GeoUDF independently on different datasets to achieve optimal performance. \Cref{qualitative_baseline} shows the qualitative comparison between our methods and baselines. To evaluate performance, we compare our approach with other baseline models in terms of $L_1$-Chamfer Distance (CD), F1-Score (setting thresholds of 0.005 and 0.01), and normal consistence (NC) metrics between the ground truth meshes and the meshes extracted from learned UDFs.
For a fair comparison, we adopt the same DCUDF~\cite{Hou2023DCUDF} method for mesh extraction. All experiments are conducted on NVIDIA RTX 4090 GPU.

\begin{table}[!t]
  \centering
  \caption{Qualitative comparison of different UDF learning methods. ``Normal'' indicates whether the method requires point cloud normals during learning. ``Feature Type''' refers to whether the information required during training is global or local. ``Noise'' and ``Outlier'' indicate whether the method can handle the presence of noise and outliers in point clouds.}
   \vspace{-0.1in}
  \resizebox{1.0\linewidth}{!}{\begin{tabular}{l c c c c c c}
    \toprule
    Methods  & Input & 
    Normal  &Learning Type & Feature Type &
    Noise   &
    Outlier \\
    \midrule
    CAP-UDF~\cite{Zhou2022CAP-UDF}  & Dense &Not required & Unsupervised & Global &\xmark &\xmark \\
    LevelSetUDF~\cite{zhou2023levelset}  & Dense &Not required & Unsupervised &Global  &\textcolor{red}{\cmark} &\xmark \\
    DUDF~\cite{fainstein2024dudf} & Dense &Required & Unsupervised &Global  &\xmark &\xmark \\
    GeoUDF~\cite{ren2023geoudf} & Sparse  &Not required &Supervised &Local  &\xmark &\xmark \\
    \midrule
    Ours  & Dense&Not required & Supervised &Local &\textcolor{red}{\cmark} & \textcolor{red}{\cmark} \\
  \bottomrule
  \end{tabular}}
    \vspace{-0.09in}
  \label{qualitative_baseline}
  \vspace{-0.18in}
\end{table}

\begin{figure*}[!t]
    \centering
     \scriptsize
    \rotatebox{90}{Input(48K)}
    \hspace{0.02in}
    \includegraphics[height=0.7in]{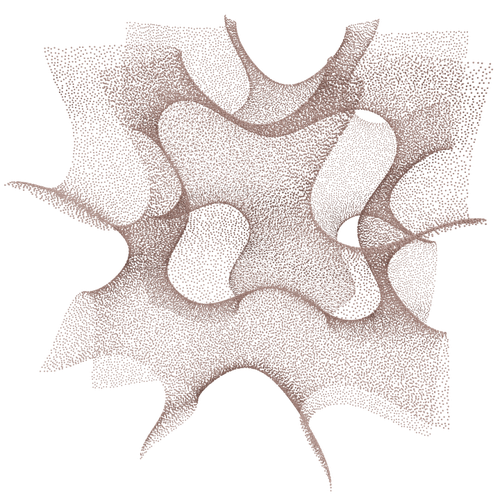}
    \hspace{0.04in}
    \includegraphics[height=0.7in]{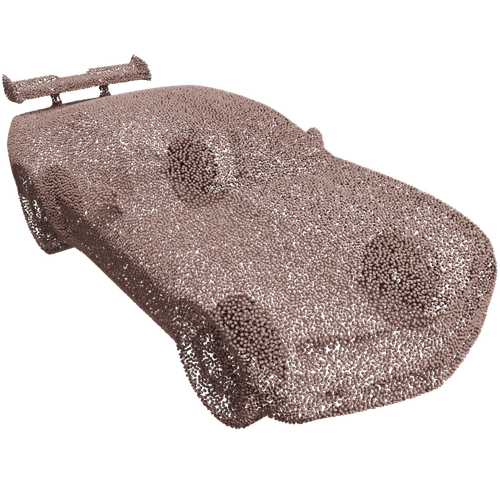}
    \hspace{0.04in}
    \includegraphics[height=0.7in]{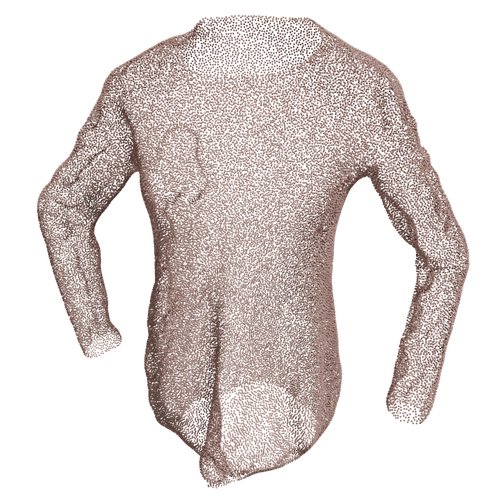}
    \hspace{0.04in}
    \includegraphics[height=0.7in]{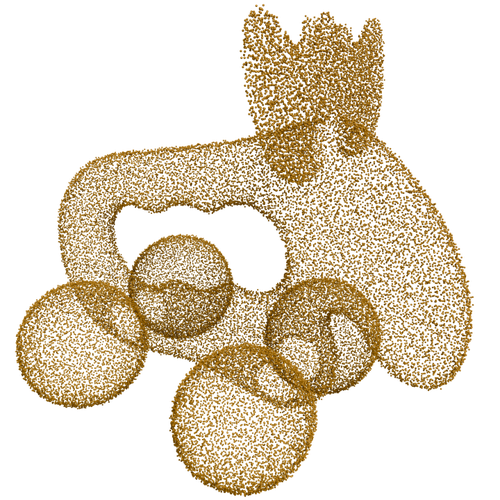}
    \hspace{0.04in}
    \includegraphics[height=0.7in]{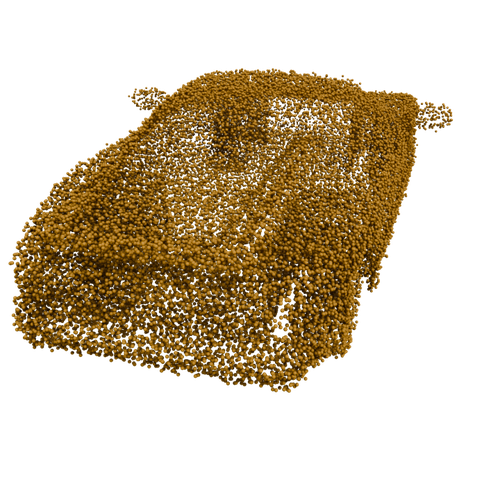}
    \hspace{0.04in}
    \includegraphics[height=0.7in]{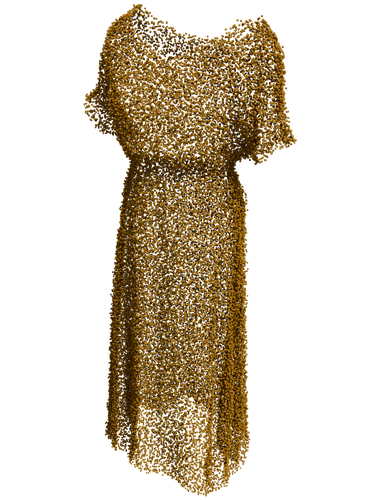}
    \hspace{0.04in}
    \includegraphics[height=0.7in]{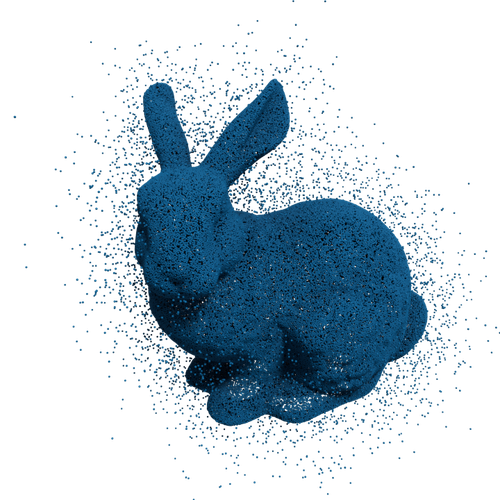}
    \hspace{0.04in}
    \includegraphics[height=0.7in]{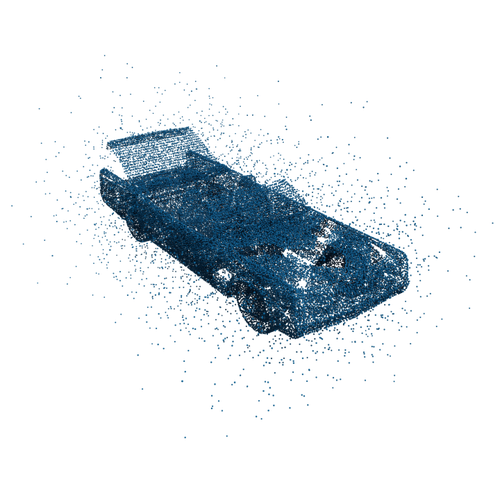}
    \hspace{0.04in}
    \includegraphics[height=0.7in]{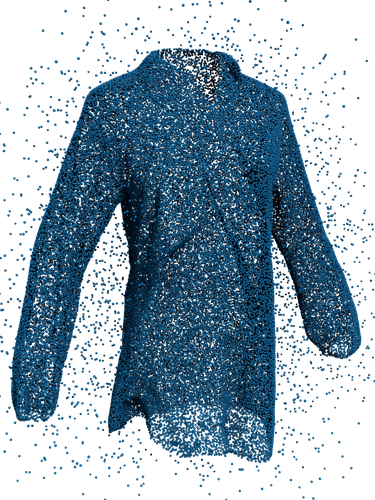}
    \\
    \rotatebox{90}{GT}
    \hspace{0.02in}
    \includegraphics[height=0.7in]{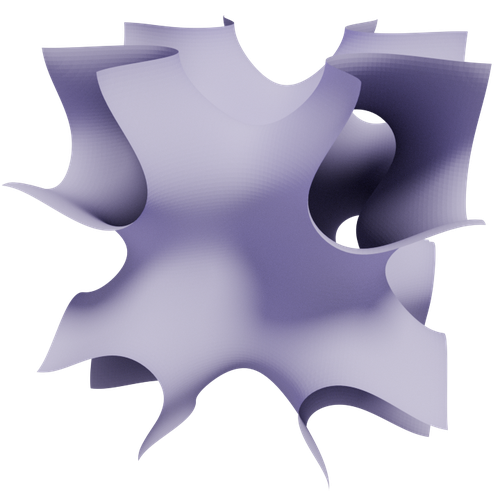}
    \hspace{0.04in}
    \includegraphics[height=0.7in]{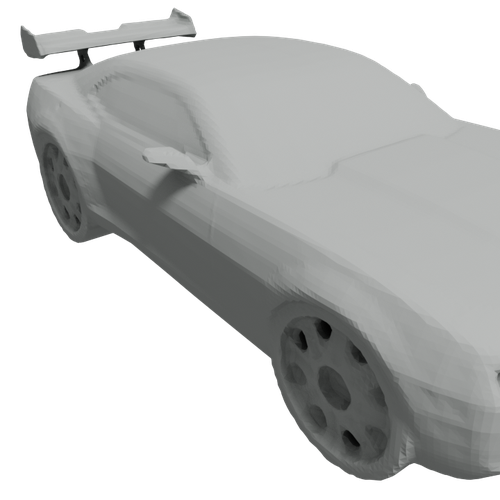}
    \hspace{0.04in}
    \includegraphics[height=0.7in]{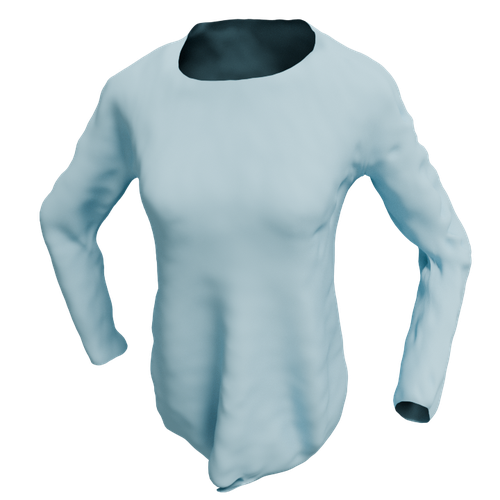}
    \hspace{0.04in}
    \includegraphics[height=0.7in]{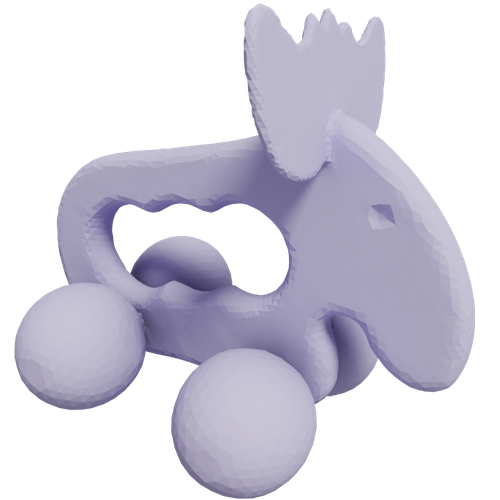}
    \hspace{0.04in}
    \includegraphics[height=0.7in]{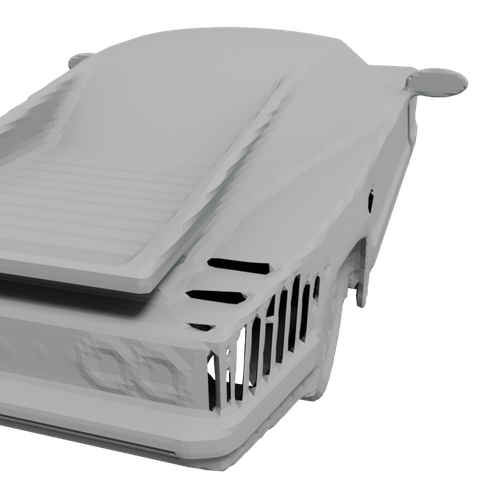}
    \hspace{0.04in}
    \includegraphics[height=0.7in]{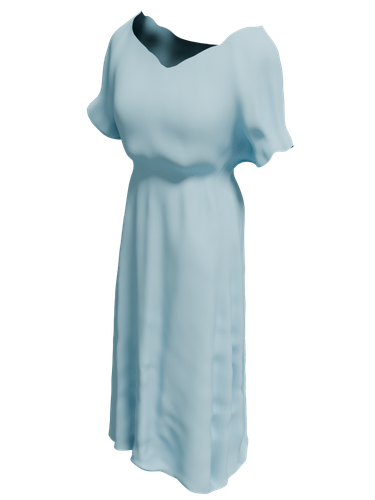}
    \hspace{0.04in}
    \includegraphics[height=0.7in]{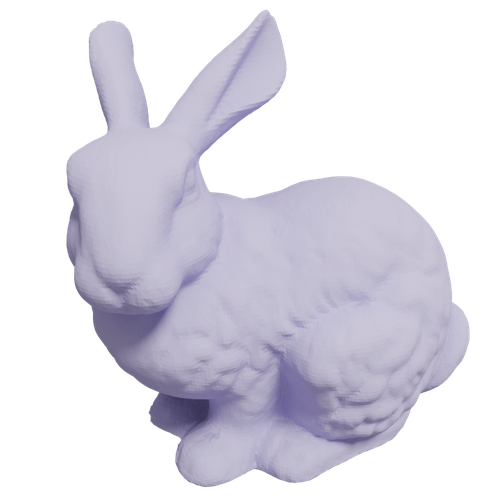}
    \hspace{0.04in}
    \includegraphics[height=0.7in]{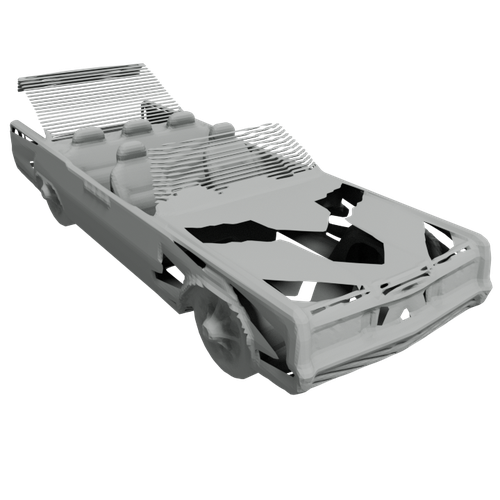}
    \hspace{0.04in}
    \includegraphics[height=0.7in]{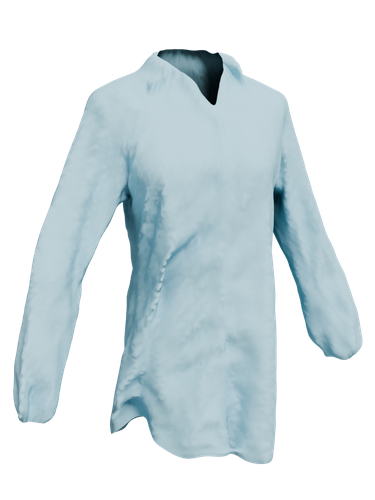}
    \\
    \rotatebox{90}{CAP-UDF}
    \hspace{0.02in}
     \includegraphics[height=0.7in]{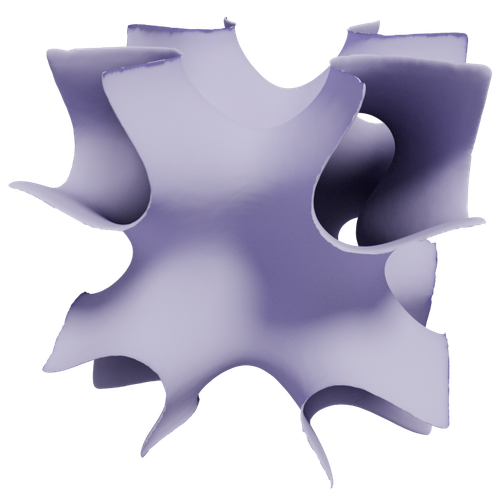}
    \hspace{0.04in}
    \includegraphics[height=0.7in]{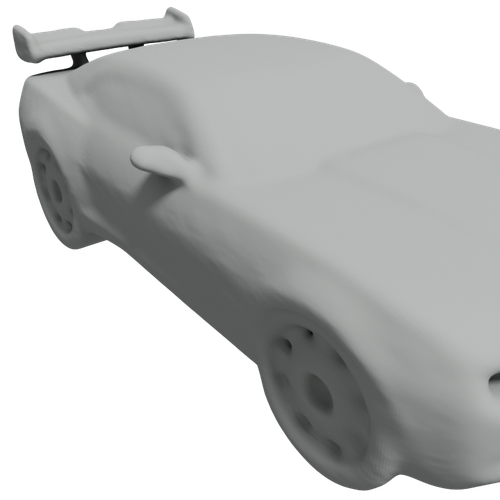}
    \hspace{0.04in}
    \includegraphics[height=0.7in]{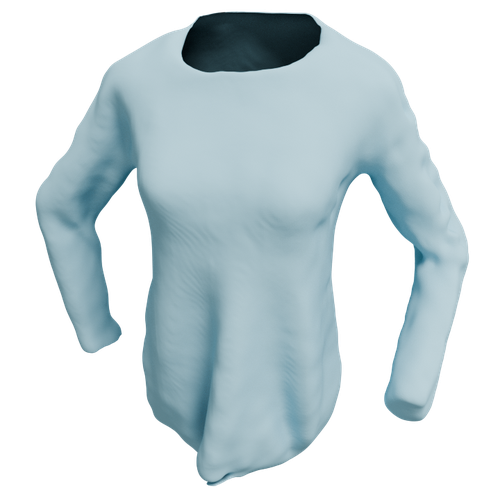}
    \hspace{0.04in}
    \includegraphics[height=0.7in]{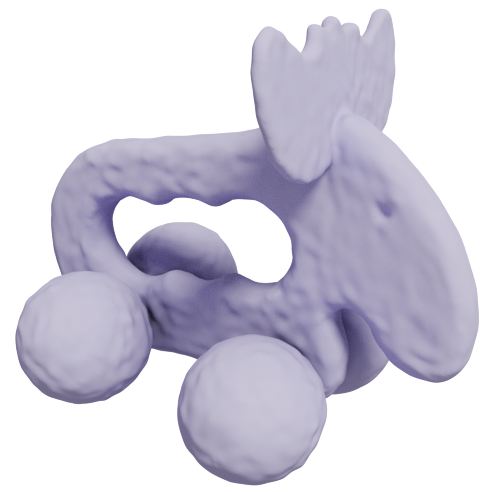}
    \hspace{0.04in}
    \includegraphics[height=0.7in]{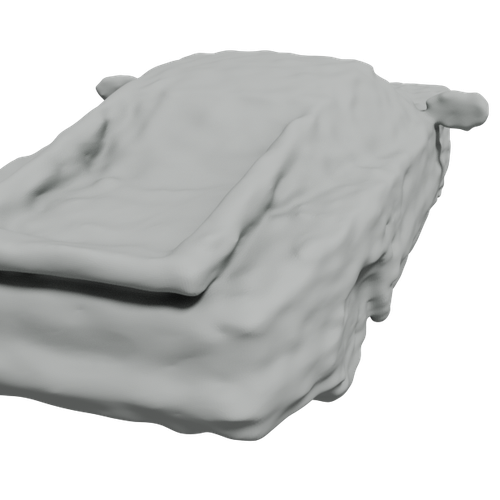}
    \hspace{0.04in}
    \includegraphics[height=0.7in,trim=1in 0 1in 0]{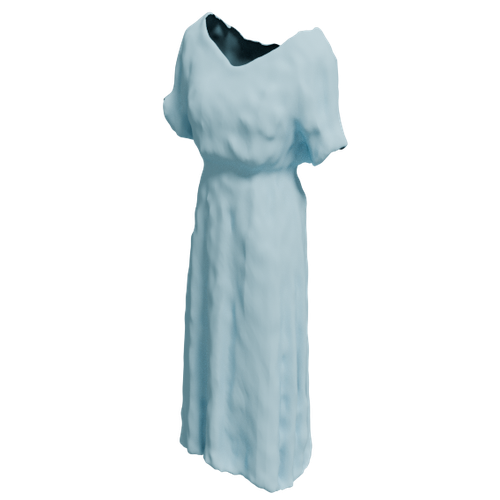}
    \hspace{0.04in}
    \includegraphics[height=0.7in]{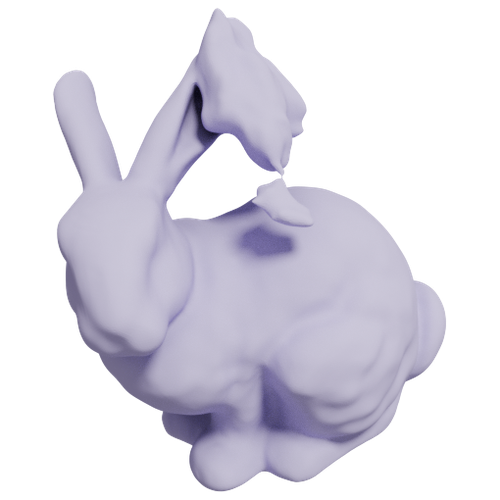}
    \hspace{0.04in}
    \includegraphics[height=0.7in]{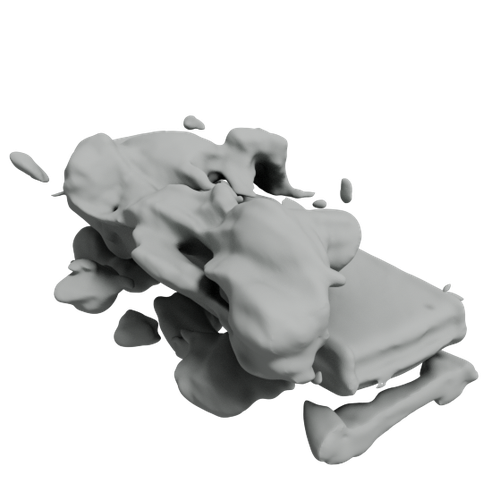}
    \hspace{0.04in}
    \includegraphics[height=0.7in,trim=1in 0 1in 0]{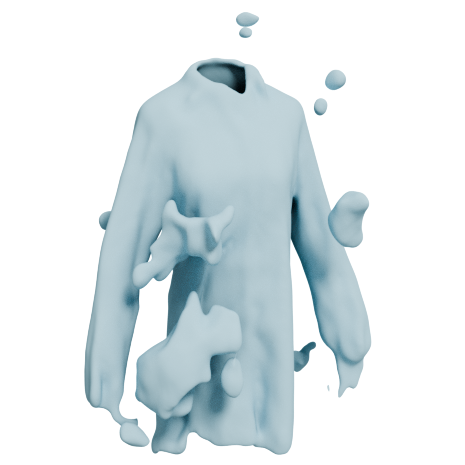}
    \\
    \rotatebox{90}{LevelSetUDF}\hspace{0.02in}
    \includegraphics[height=0.7in]{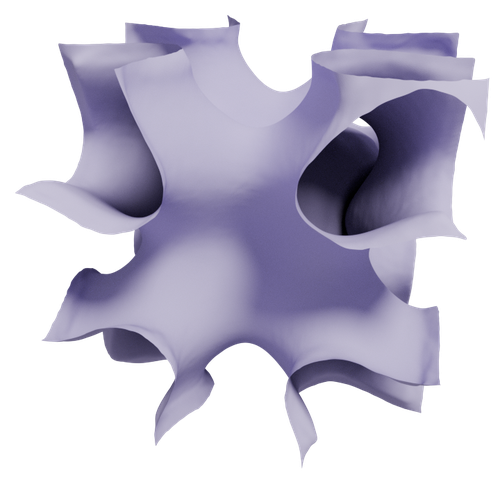}
    \hspace{0.04in}
    \includegraphics[height=0.7in]{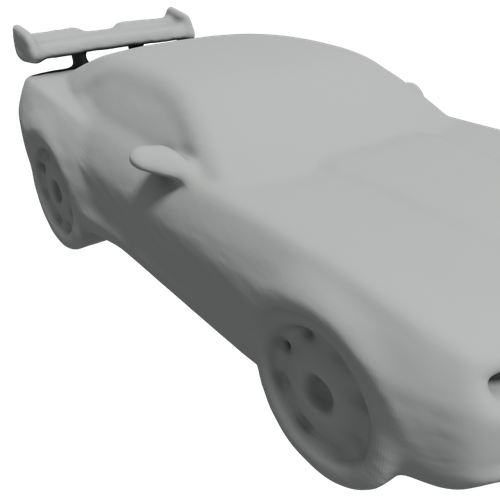}
    \hspace{0.04in}
    \includegraphics[height=0.7in]{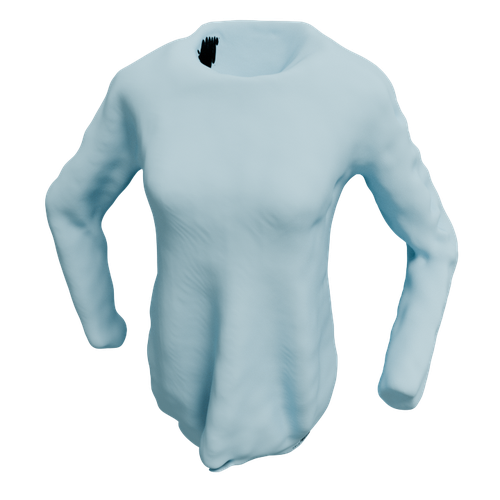}
    \hspace{0.04in}
    \includegraphics[height=0.7in]{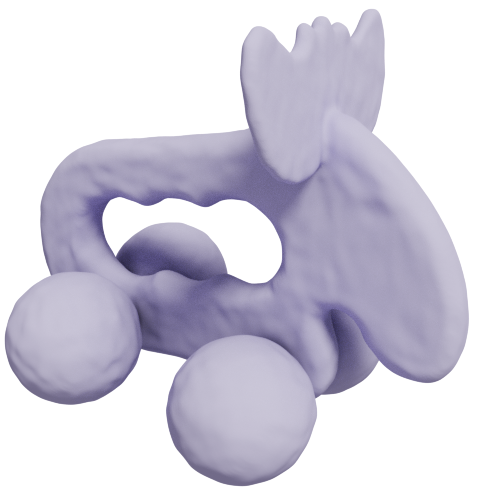}
    \hspace{0.04in}
    \includegraphics[height=0.7in]{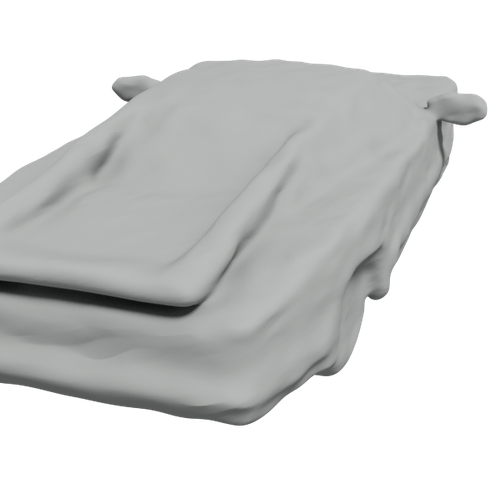}
    \hspace{0.04in}
    \includegraphics[height=0.7in,trim=1in 0 1in 0]{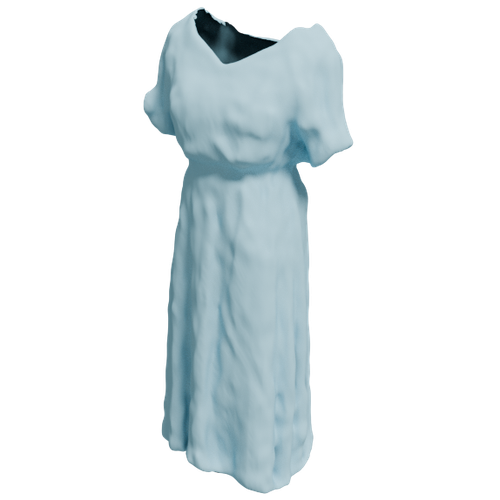}
    \hspace{0.04in}
    \includegraphics[height=0.7in]{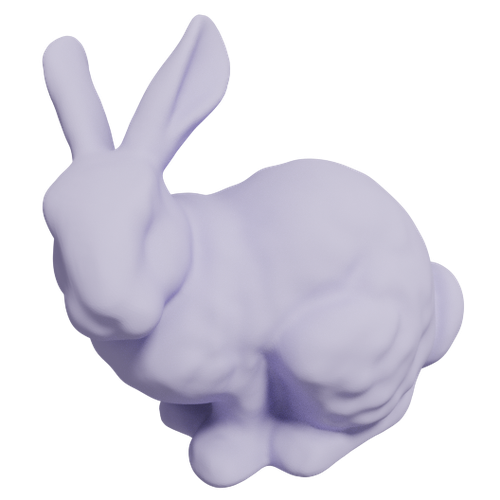}
    \hspace{0.04in}
    \includegraphics[height=0.7in]{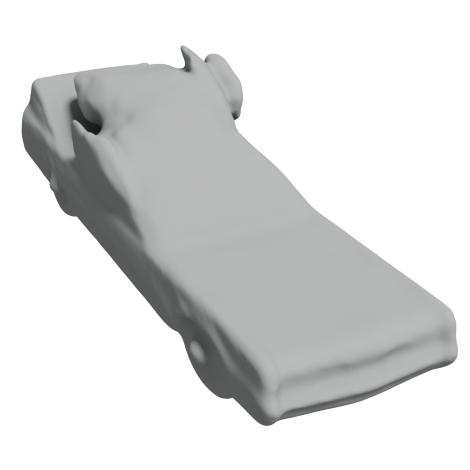}
    \hspace{0.04in}
    \includegraphics[height=0.7in,trim=1in 0 1in 0]{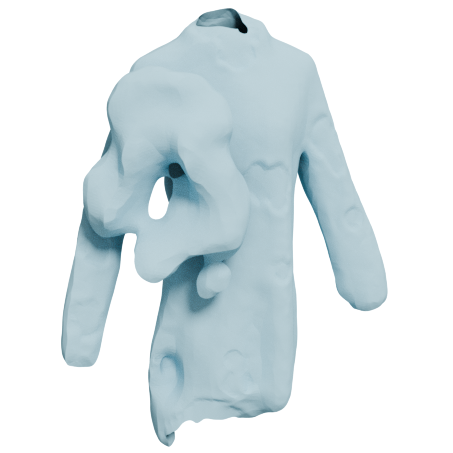}
    \\
    \rotatebox{90}{GeoUDF}\hspace{0.02in}
    \includegraphics[height=0.7in]{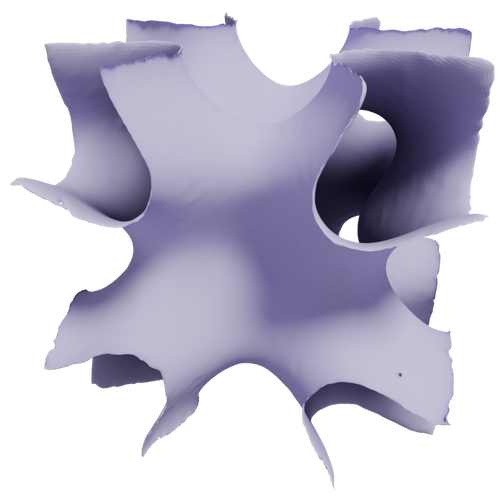}
    \hspace{0.04in}
    \includegraphics[height=0.7in]{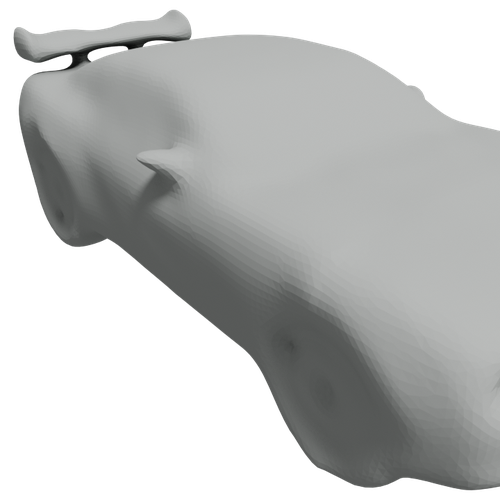}
    \hspace{0.04in}
    \includegraphics[height=0.7in]{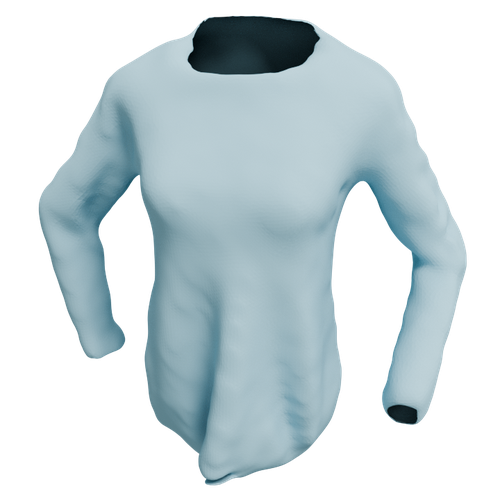}
    \hspace{0.04in}
    \includegraphics[height=0.7in]{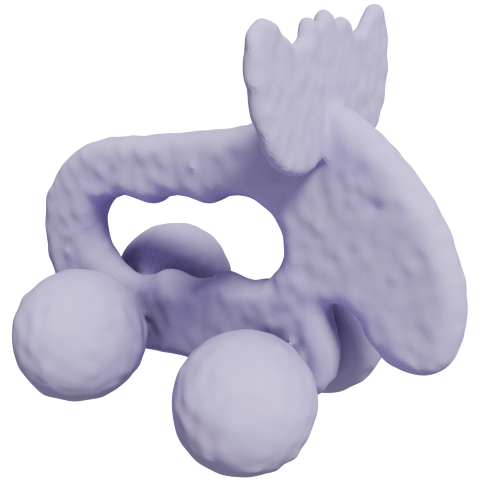}
    \hspace{0.04in}
    \includegraphics[height=0.7in]{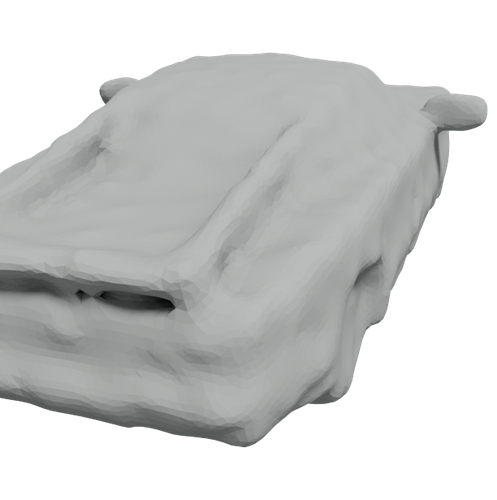}
    \hspace{0.04in}
    \includegraphics[height=0.7in,trim=1in 0 1in 0]{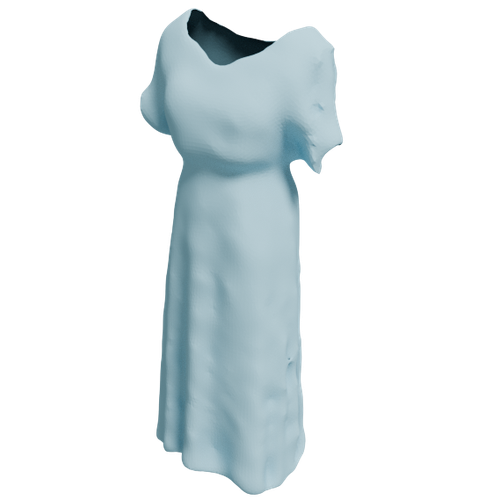}
    \hspace{0.04in}
    \includegraphics[height=0.7in]{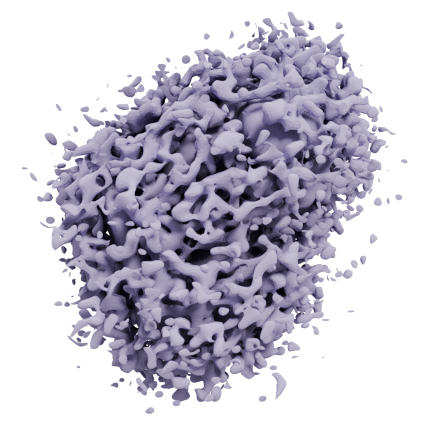}
    \hspace{0.04in}
    \includegraphics[height=0.7in]{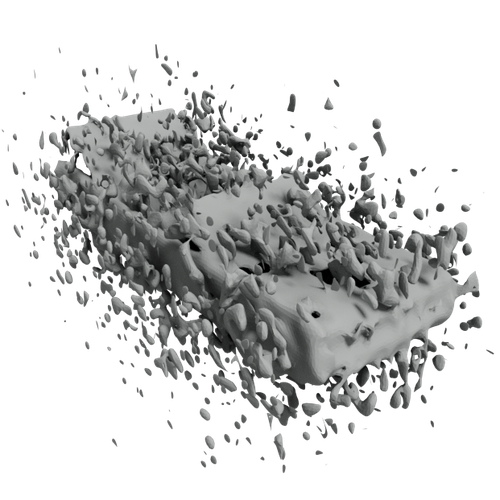}
    \hspace{0.04in}
    \includegraphics[height=0.7in,trim=1in 0 1in 0]{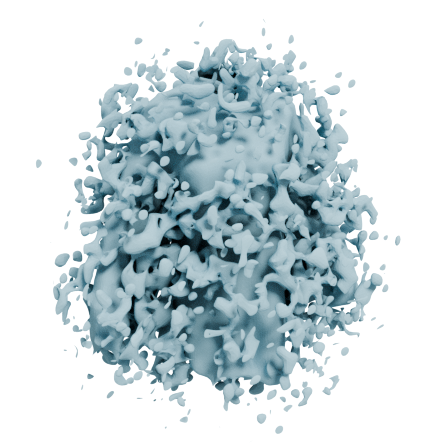}
    \\
    \rotatebox{90}{DUDF}\hspace{0.02in}
    \includegraphics[height=0.7in]{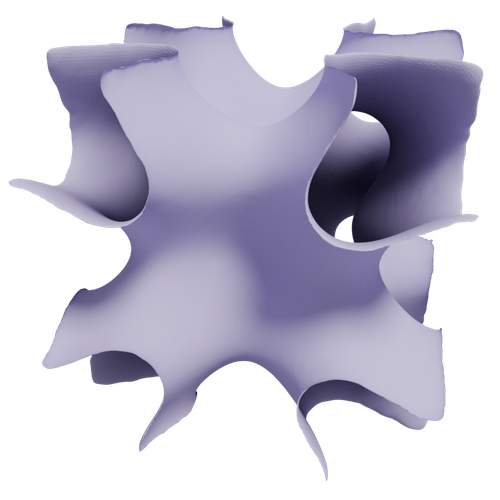}
    \hspace{0.04in}
    \includegraphics[height=0.7in]{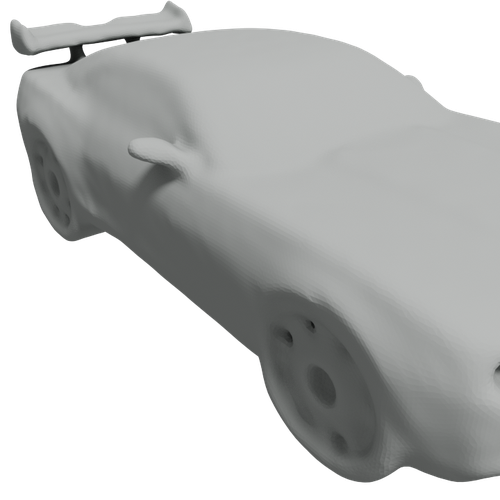}
    \hspace{0.04in}
    \includegraphics[height=0.7in]{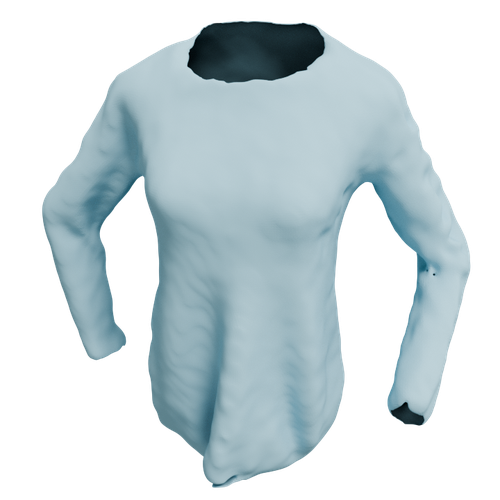}
    \hspace{0.04in}
    \includegraphics[height=0.7in]{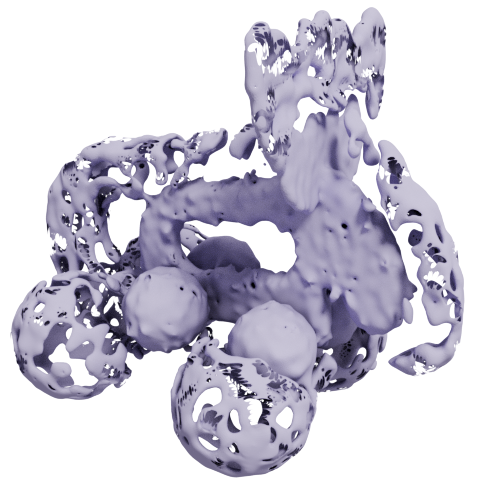}
    \hspace{0.04in}
    \includegraphics[height=0.7in]{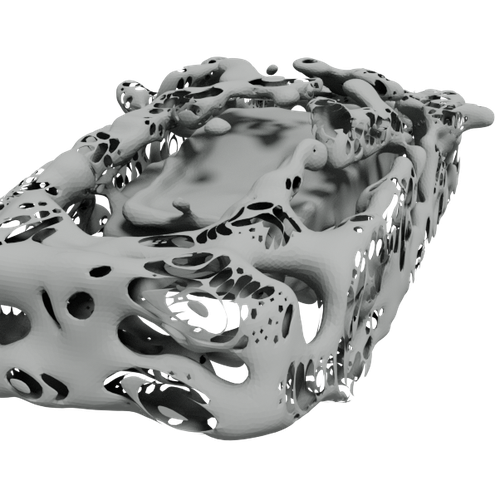}
    \hspace{0.04in}
    \includegraphics[height=0.7in,trim=1in 0 1in 0]{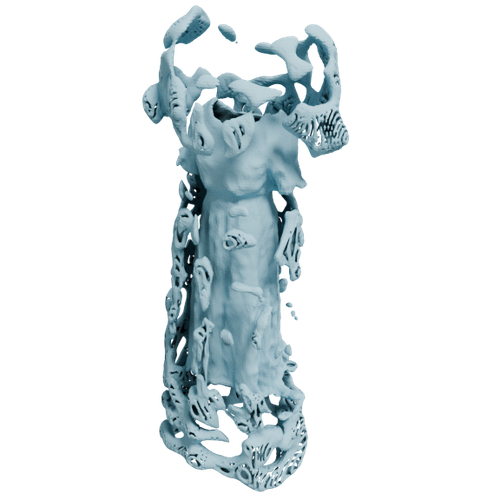}
    \hspace{0.04in}
    \includegraphics[height=0.7in]{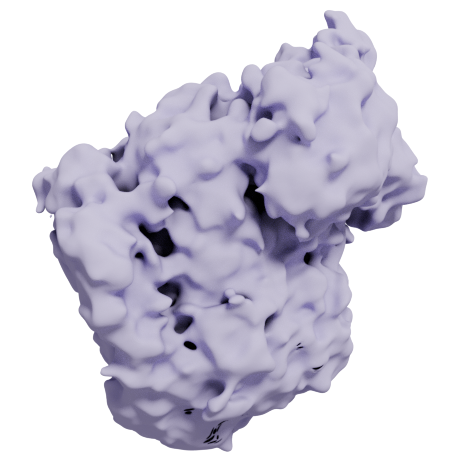}
    \hspace{0.04in}
    \includegraphics[height=0.7in]{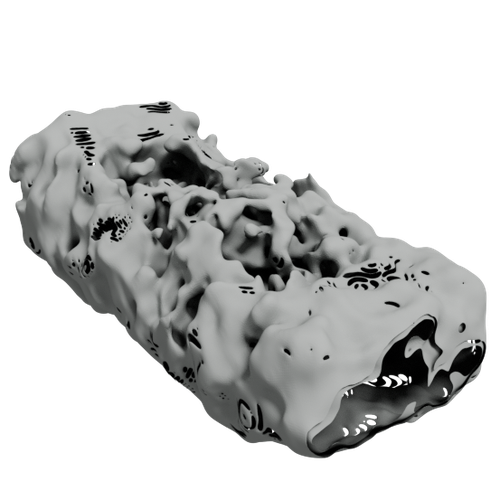}
    \hspace{0.04in}
    \includegraphics[height=0.7in,trim=1in 0 1in 0]{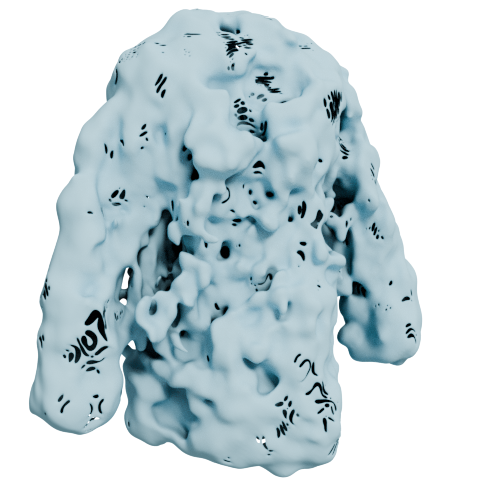}
    \\
    \rotatebox{90}{Ours}\hspace{0.02in}
    \includegraphics[height=0.7in]{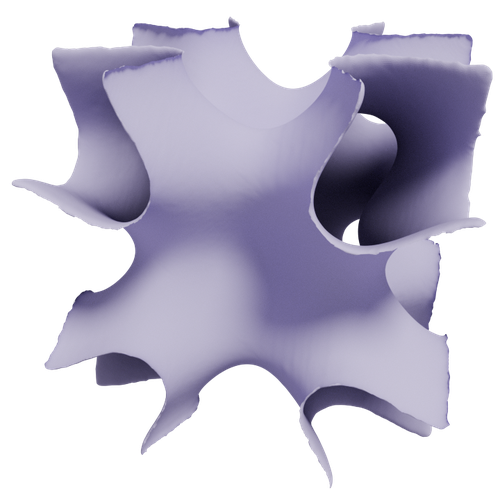}
    \hspace{0.04in}
    \includegraphics[height=0.7in]{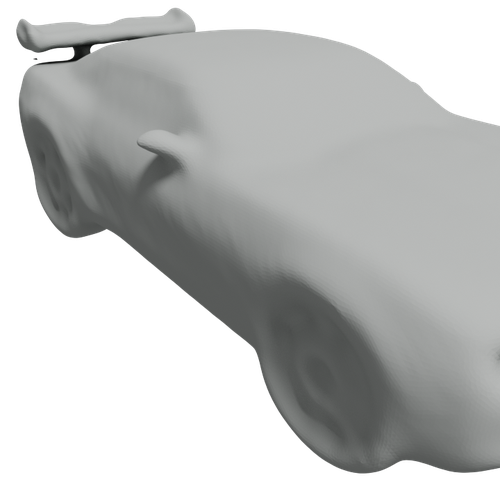}
    \hspace{0.04in}
    \includegraphics[height=0.7in]{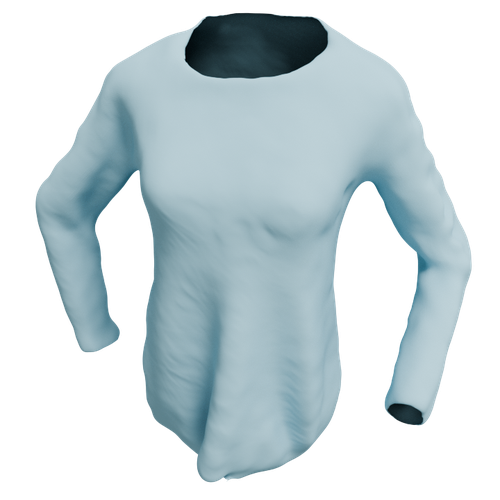}
    \hspace{0.04in}
    \includegraphics[height=0.7in]{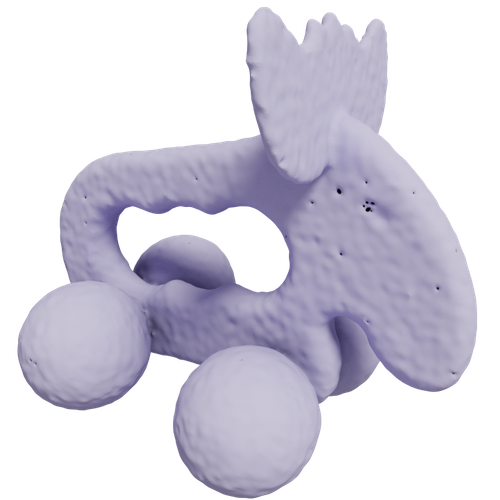}
    \hspace{0.04in}
    \includegraphics[height=0.7in]{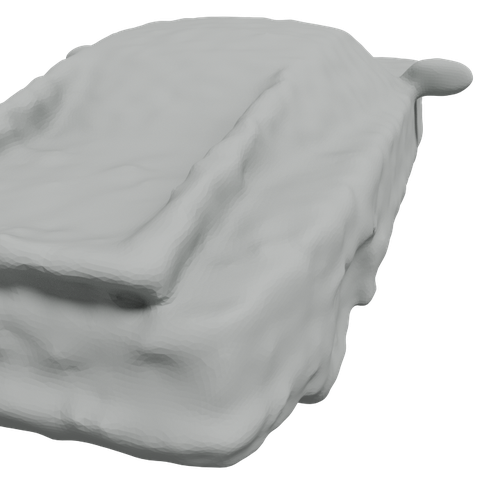}
    \hspace{0.04in}
    \includegraphics[height=0.7in]{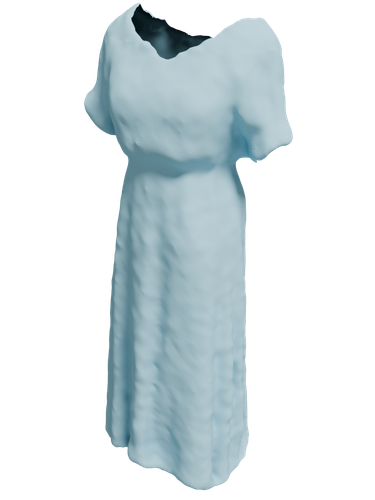}
    \hspace{0.04in}
    \includegraphics[height=0.7in]{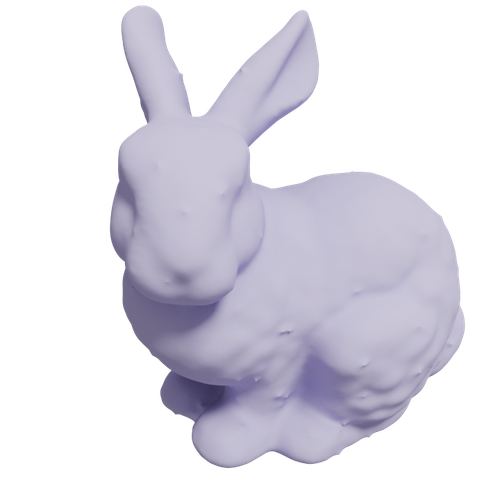}
    \hspace{0.04in}
    \includegraphics[height=0.7in]{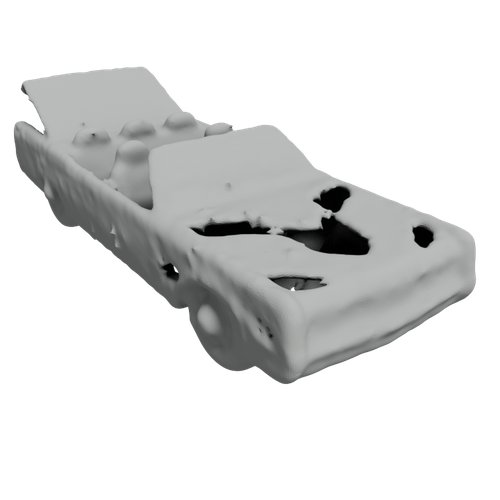}
    \hspace{0.04in}
    \includegraphics[height=0.7in]{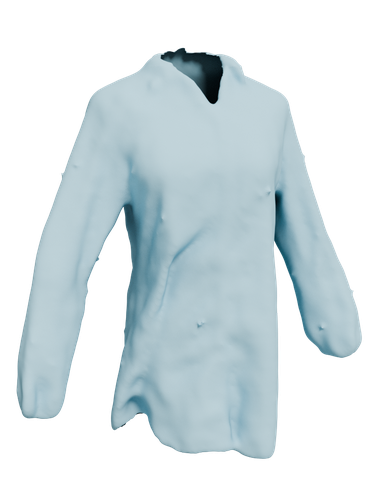}
    \\
    \hspace{0.1in}
    \rule{2.3in}{0.6pt}\hspace{0.04in}
    \rule{2.2in}{0.6pt}\hspace{0.04in}
    \rule{2.1in}{0.6pt}\\
    \hspace{0.1in}
    \makebox[0.32\textwidth]{Clean}\hfill
    \makebox[0.32\textwidth]{Noise(0.25\%)}\hfill
    \makebox[0.32\textwidth]{Outliers(10\%)}
    \vspace{-0.1in}
    \caption{Visual comparisons of reconstruction results on the synthetic dataset.  We provide more results in the supplementary materials.}
    \vspace{-0.2in}
    \label{fig:visual-results}
\end{figure*}

\begin{figure}[b]
\centering
\includegraphics[width=0.8\linewidth]{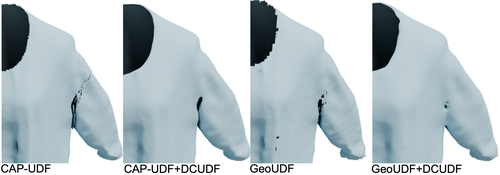}
\vspace{-0.1in}
\caption{We compare other methods using their own mesh extraction techniques against the DCUDF approach.}
\label{fig:compare-extraction}
\end{figure}

\subsection{Experimental results}
\textbf{Synthetic data.} For general 3D graphic models, ShapeNetCars, and DeepFashion3D, we obtain dense point clouds by randomly sampling on meshes.
Considering that GeoUDF~\cite{ren2023geoudf} is a supervised method, we retrain it on ShapeNetCars, and DeepFashion3D, which are randomly partitioned into training (70\%), testing (20\%), and validation subsets (10\%). 
All models are evaluated in the validation sets, which remain unseen by any of the UDF learning models prior to evaluation.  \Cref{fig:visual-results} illustrates the visual comparison of reconstruction results, and \Cref{tab:quantitative} presents the quantitative comparison in terms of evaluation metrics. 
We test each method using their own mesh extraction technique, as shown in \cref{fig:compare-extraction}, which display obvious visual artifacts such as small holes and non-smoothness. We thus apply DCUDF~\cite{Hou2023DCUDF} , the state-of-art method, to each baseline model , extracting the surfaces as significantly higher quality meshes. Since our method utilizes DCUDF for surface extraction, we adopt it as the default technique to ensure consistency and fairness in comparisons with the baselines.
Our method achieves stable results in reconstructing various types of surfaces, including both open and closed surfaces, and exhibits performance comparable to that of the SOTA methods. Noting that DUDF~\cite{fainstein2024dudf} requires normals during training, and GeoUDF utilizes the KNN approach to determine the nearest neighbors of the query points. As a result,  DUDF and GeoUDF are less stable when dealing with point clouds with noise and outliers, as shown in \cref{fig:visual-results}.

\begin{table*}[!t]
    \centering
    \begin{minipage}[t]{0.64\textwidth}
    \centering
    \caption{We compare our method with other UDF learning methods in terms of $L_1$-Chamfer distance ($\times 100$), F-score with thresholds of 0.005 and 0.01, and normal consistence. The best results are highlighted with \colorbox{tabfirst}{1st} and \colorbox{tabsecond}{2nd}.} 
    \vspace{-0.15in}
    \label{tab:quantitative}
    \resizebox{1.0\textwidth}{!}{
    \begin{tabular}{c|l|c|cc|c|c|cc|c|c|cc|c}
    \toprule
        & & \multicolumn{4}{c|}{Clean} & \multicolumn{4}{c|}{Noise} & \multicolumn{4}{c}{Outliers} \\ \cmidrule{3-14}
        & Method      &CD $\downarrow$  & $F1^{0.005}$   & $F1^{0.01}$  &NC $\uparrow$ &CD $\downarrow$ & $F1^{0.005}$   & $F1^{0.01}$ &NC $\uparrow$  &CD $\downarrow$ & $F1^{0.005}$   & $F1^{0.01}$&NC $\uparrow$ \\ \cmidrule{1-14}
        
        \multirow{6}{*}{\rotatebox{90}{\scriptsize ShapeNetCars \cite{shapenet2015}}} 
        & \small CAP-UDF \cite{Zhou2022CAP-UDF} &1.324 &0.451 &0.884 &0.945 &1.208 &0.286 &0.632 &0.833 &2.918 &0.194 &0.304 &0.725\\
        & \small LevelSetUDF \cite{zhou2023levelset} &1.223 &0.464 &0.892 &\cellcolor{tabsecond}0.950 &\cellcolor{tabsecond}1.176 &\cellcolor{tabsecond}0.299 &\cellcolor{tabsecond}0.654 &0.842 &2.763 &\cellcolor{tabsecond}0.200 &\cellcolor{tabsecond}0.379 &0.753\\
        & \small GeoUDF  \cite{ren2023geoudf} &0.832 &0.562 &\cellcolor{tabsecond}0.901 &0.947 &1.869 &0.205 &0.608 &\cellcolor{tabsecond}0.847 &\cellcolor{tabsecond}2.598 &0.144 &0.327 &\cellcolor{tabsecond}0.757\\
        & \small DUDF \cite{fainstein2024dudf}  &\cellcolor{tabsecond}0.773 &\cellcolor{tabfirst}0.697 &0.871 &0.945 &1.939 &0.214 &0.508 &0.826 &2.949 &0.135 &0.329 &0.722\\ 
        &\small \textbf{Ours}   &\cellcolor{tabfirst}0.611 &\cellcolor{tabsecond}0.645 &\cellcolor{tabfirst}0.927 &\cellcolor{tabfirst}0.967 &\cellcolor{tabfirst}0.614 &\cellcolor{tabfirst}0.546 &\cellcolor{tabfirst}0.913 &\cellcolor{tabfirst}0.966 &\cellcolor{tabfirst}0.705 &\cellcolor{tabfirst}0.437 &\cellcolor{tabfirst}0.751 &\cellcolor{tabfirst}0.944\\ \cmidrule{1-14}
        
        \multirow{6}{*}{\rotatebox{90}{\tiny DeepFashion3D \cite{liuLQWTcvpr16DeepFashion}}} 
        & \small CAP-UDF \cite{Zhou2022CAP-UDF} &0.885 &0.353 &0.809 &0.973 &0.803 &0.375 &0.719 &0.871 &5.150 &0.173 &0.425 &0.863\\
        & \small LevelSetUDF \cite{zhou2023levelset} &0.881 &0.335 &0.848 &\cellcolor{tabsecond}0.983 &0.723 &0.384 &0.750 &0.874 &5.083 &\cellcolor{tabsecond}0.278 &\cellcolor{tabsecond} 0.430 &\cellcolor{tabsecond} 0.882\\
        & \small GeoUDF  \cite{ren2023geoudf} &0.625 &0.525 &0.957 &0.973 &0.676 &\cellcolor{tabsecond}0.506 &\cellcolor{tabsecond}0.951 &\cellcolor{tabsecond}0.882 &3.883 &0.146 &0.310 &0.764\\
        & \small DUDF \cite{fainstein2024dudf}  &\cellcolor{tabfirst} 0.506 &\cellcolor{tabfirst}0.586 &\cellcolor{tabfirst}0.972 &0.965 &0.934 &0.333 &0.530 &0.849 &\cellcolor{tabsecond} 2.558 &0.118 &0.265 &0.821\\
        &\textbf{Ours} &\cellcolor{tabsecond}0.548 &\cellcolor{tabsecond}0.565 &\cellcolor{tabsecond}0.966 &\cellcolor{tabfirst}0.988 &\cellcolor{tabfirst}0.614 &\cellcolor{tabfirst}0.541 &\cellcolor{tabfirst}0.967 &\cellcolor{tabfirst}0.906 &\cellcolor{tabfirst} 0.559 &\cellcolor{tabfirst} 0.463 &\cellcolor{tabfirst} 0.966 &\cellcolor{tabfirst} 0.909\\ \cmidrule{1-14}
    \end{tabular}}  
    \end{minipage}
    \hfill
    \begin{minipage}[t]{0.34\textwidth}
        \caption{Comparison of time efficiency. We measured the average runtime in minutes. "\#Params" denotes the number of network parameters, while "Size" refers to the storage space occupied by these parameters.  }
        \label{tab:time}
        \resizebox{1.0\linewidth}{!}{
        \begin{tabular}{l|c|c|c|c|c}
            \toprule
            Method  & SRB & DeepFashion3D & ShapeNetCars & \#Param & Size (KB) \\
            \midrule
            CAP-UDF  & 15.87 & 10.5 & 10.6 & 463100  & 1809\\
            \textbf{LoSF + CAP-UDF} & \textbf{6.84} & \textbf{4.32} & \textbf{4.40} & -  & -\\
            \midrule
            LevelSetUDF  & 15.08 & 13.65 & 14.67 & 463100 & 1809\\
            \textbf{LoSF + LevelSetUDF} & \textbf{6.13} & \textbf{4.85} & \textbf{4.97} & - & -\\
           \midrule
            DUDF  & 14.28 & 11.12 & 14.58 & 461825 & 1804\\
           \midrule
            GeoUDF  & 0.08 & 0.07 & 0.07 & 253378 & 990\\
            \midrule
            \textbf{Ours}  & \textbf{0.87} &\textbf{0.51} & \textbf{0.42} & \textbf{167127} & \textbf{653}\\
            \bottomrule
        \end{tabular}}
    \end{minipage}
    \vspace{-0.2in}
\end{table*}

\noindent\textbf{Noise \& outliers.} To evaluate our model with noisy inputs, we added Gaussian noise $\mathcal{N}(0, 0.25\%)$ to the clean data across all datasets for testing.
The middle three columns in \cref{fig:visual-results} display the reconstructed surface results from noisy point clouds, and \Cref{tab:quantitative} also presents the quantitative comparisons. It can be observed that our method can robustly reconstruct smooth surfaces from noisy point clouds. Additionally, we tested our method's performance with outliers by converting 10\% of the clean point cloud into outliers, as shown in the last three columns of \cref{fig:visual-results}. Experimental results demonstrate that our method can handle up to 50\% outliers while still achieving reasonable results. Even in the presence of both noise and outliers, our method maintains a high level of robustness. The corresponding results are provided in the supplementary materials.

\begin{figure}[!b]
  \centering
  \includegraphics[height=0.49in]{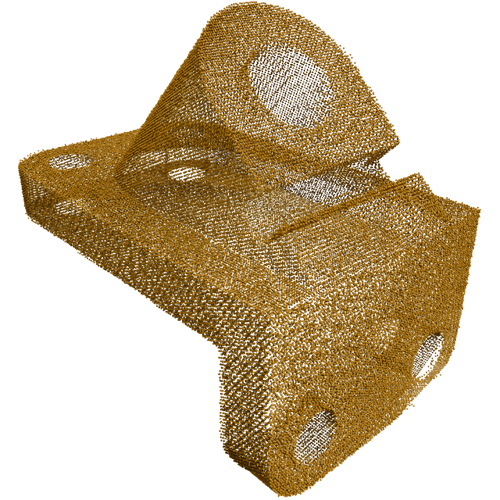}
  \hfill
  \includegraphics[height=0.49in]{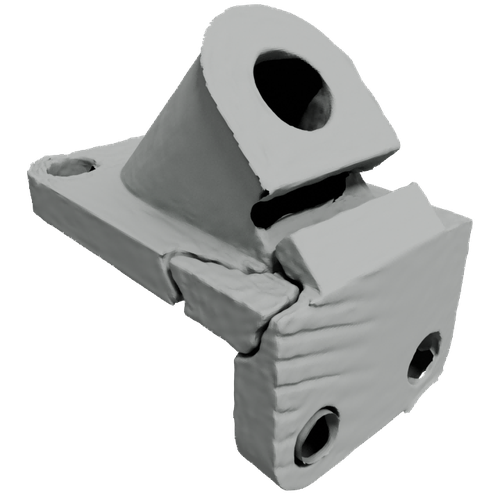}
  \hfill\hfill\hfill\hfill
  \includegraphics[height=0.49in]{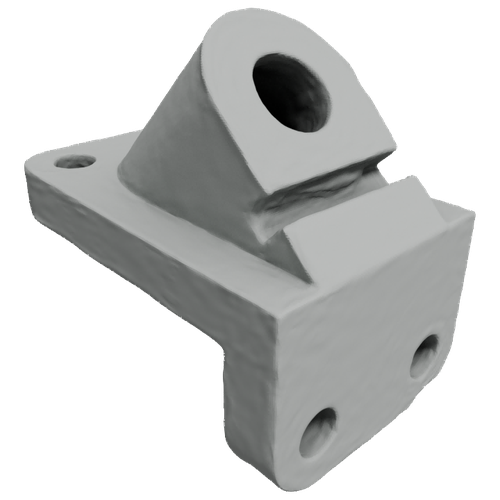}
  \hfill\hfill\hfill
  \includegraphics[height=0.49in]{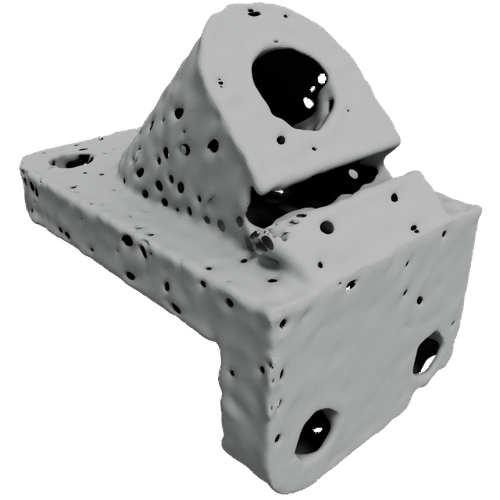}
  \hfill\hfill
  \includegraphics[height=0.49in]{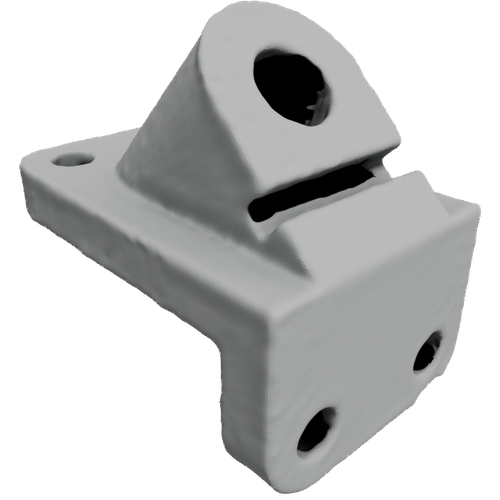}
  \hfill
  \includegraphics[height=0.49in]{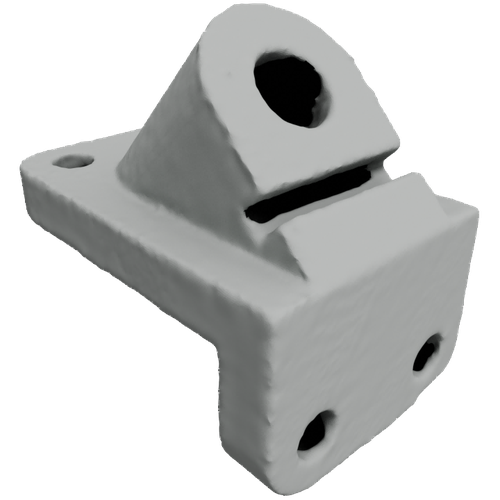}
  \\
  \vspace{-0.03in}
  \includegraphics[height=0.49in]{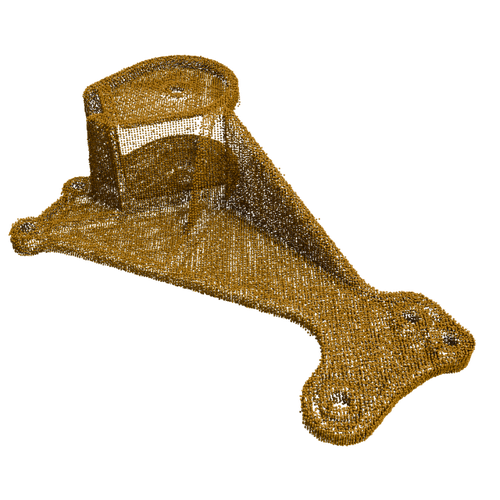}
  \hfill
  \includegraphics[height=0.49in]{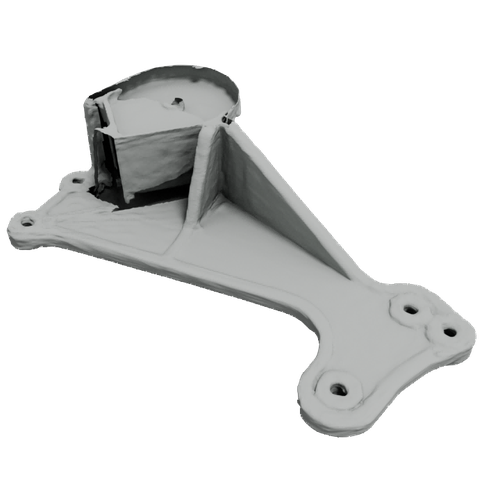}
  \hfill
  \includegraphics[height=0.49in]{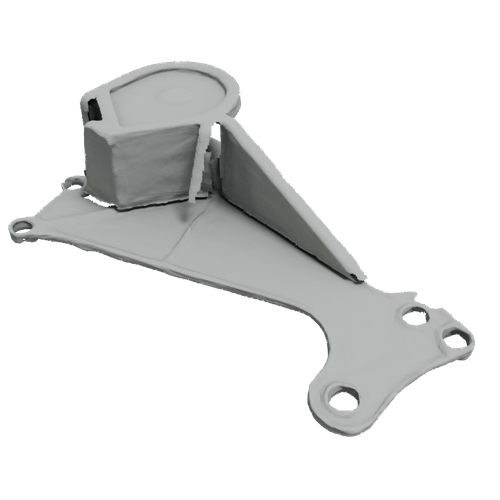}
  \hfill
  \includegraphics[height=0.49in]{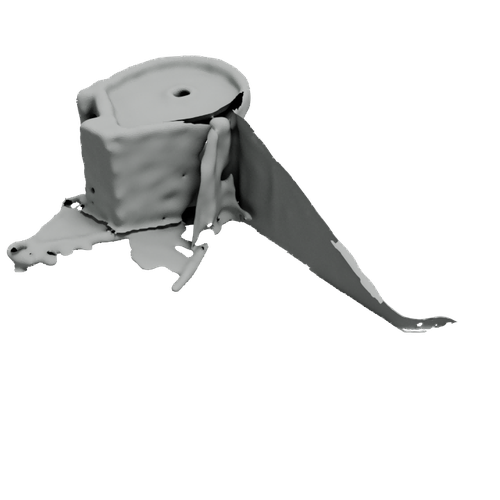}
  \hfill
  \includegraphics[height=0.49in]{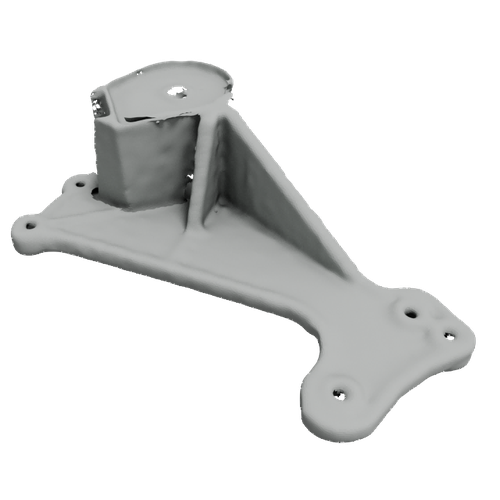}
  \hfill
  \includegraphics[height=0.49in]{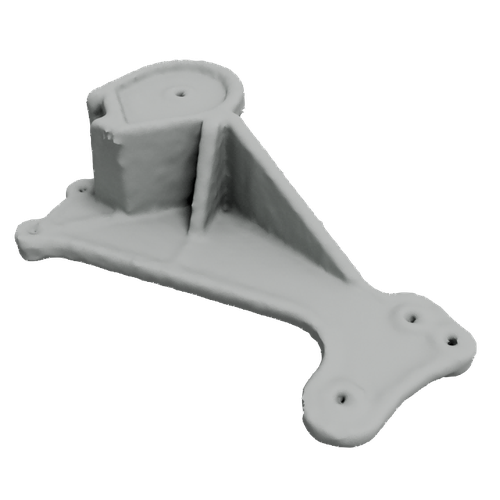}
  \\
  \vspace{-0.03in}
  \includegraphics[height=0.49in]{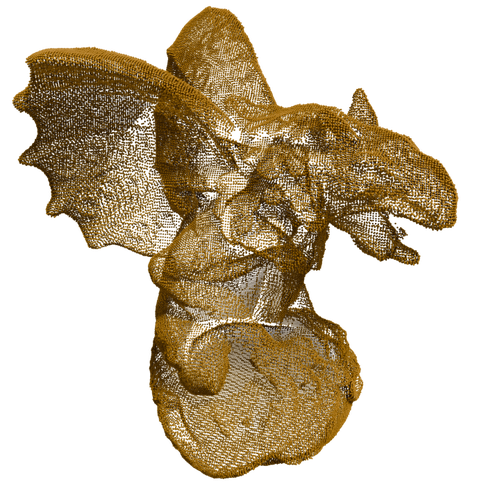}
  \hfill
  \includegraphics[height=0.49in]{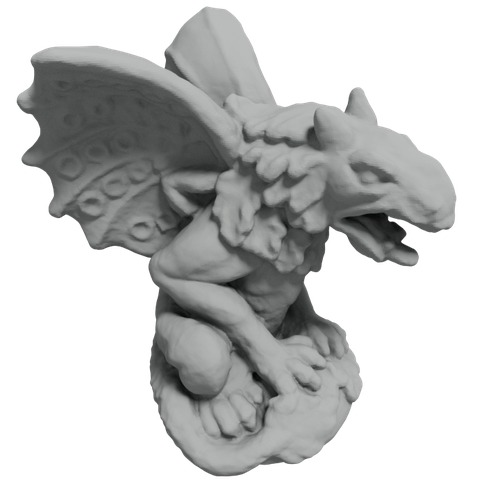}
  \hfill
  \includegraphics[height=0.49in]{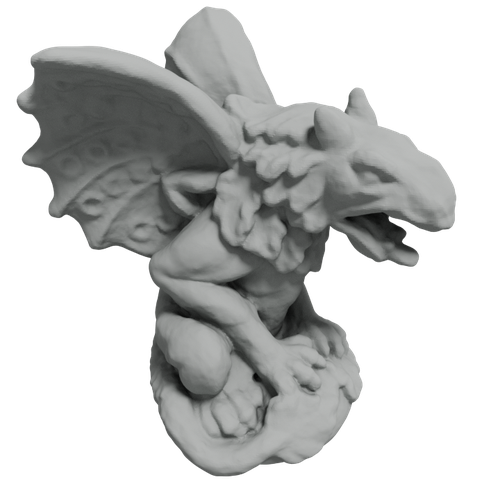}
  \hfill
  \includegraphics[height=0.49in]{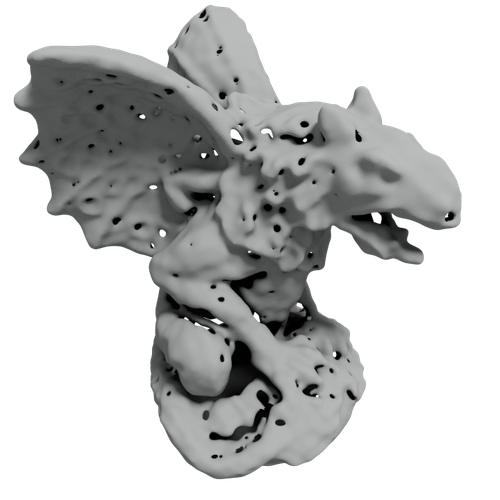}
  \hfill
  \includegraphics[height=0.49in]{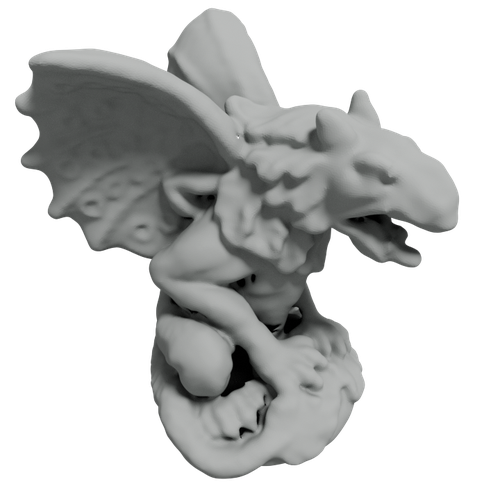}
  \hfill
  \includegraphics[height=0.49in]{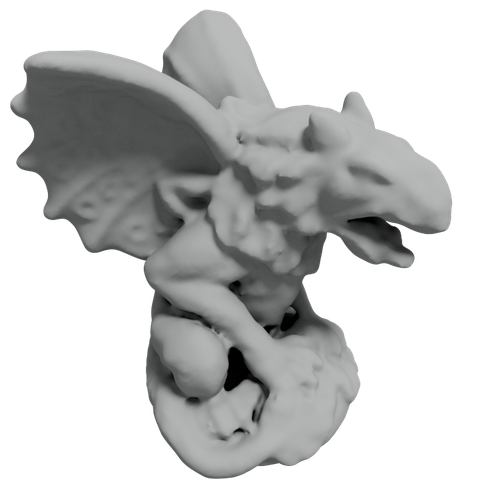}
  \\
  \vspace{-0.03in}
  \includegraphics[height=0.49in]{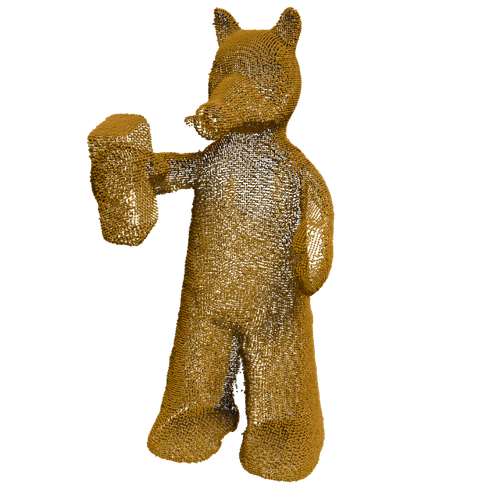}
  \hfill
  \includegraphics[height=0.49in]{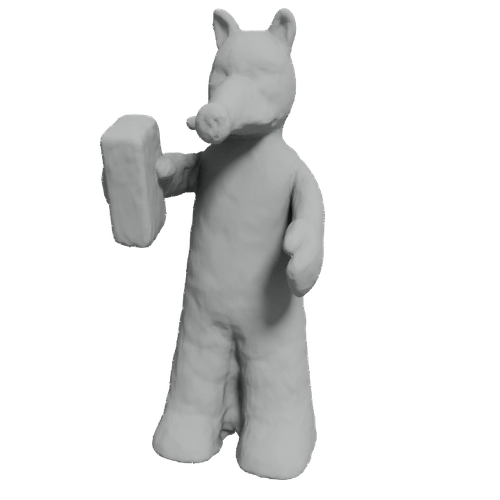}
  \hfill
  \includegraphics[height=0.49in]{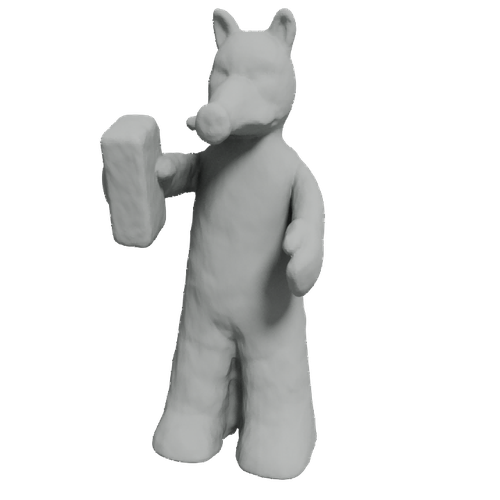}
  \hfill
  \includegraphics[height=0.49in]{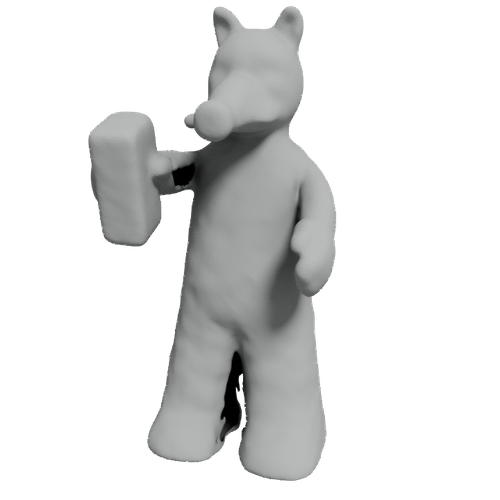}
  \hfill
  \includegraphics[height=0.49in]{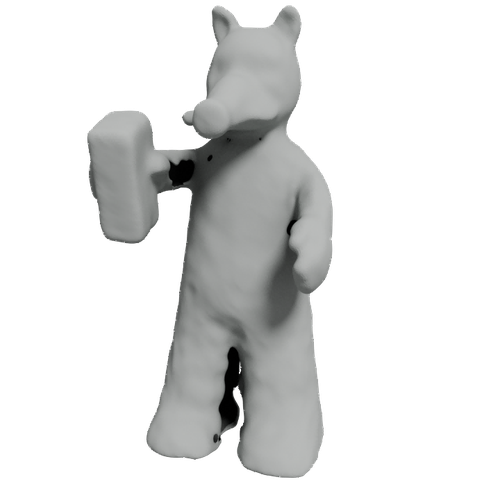}
  \hfill
  \includegraphics[height=0.49in]{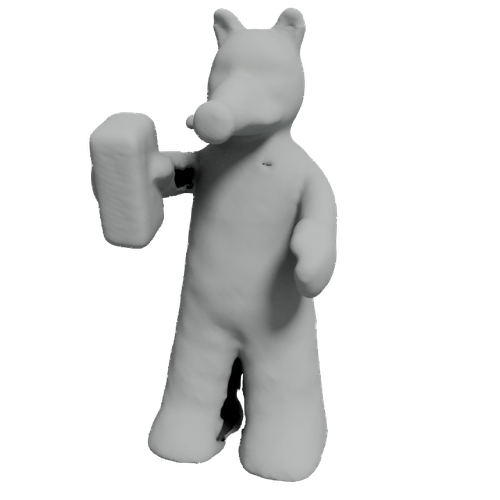}
  \\
  \vspace{-0.08in}
  \makebox[0.137\linewidth]{\scriptsize{Input}}
  \hfill
  \makebox[0.137\linewidth]{\scriptsize{CAP-UDF}}
  \hfill
  \makebox[0.137\linewidth]{\scriptsize{LevelSetUDF}}
  \hfill
  \makebox[0.137\linewidth]{\scriptsize{DUDF}}
  \hfill
  \makebox[0.137\linewidth]{\scriptsize{GeoUDF}}
  \hfill
  \makebox[0.137\linewidth]{\scriptsize{Ours}}
  \vspace{-0.1in}
    \resizebox{1.0\linewidth}{!}{
    \begin{tabular}{rccccc}
            \toprule
            Method & CAP-UDF & LevelSetUDF & DUDF & GeoUDF & Ours \\
            \midrule
            CD($\times$100) $\downarrow$ & 2.101 & 1.637 & 4.207 & 1.762 & \textbf{1.551} \\
            \midrule
            $F1^{0.005}$ $\uparrow$ & 0.410 & 0.464 & 0.378 & 0.557 & \textbf{0.589}\\
            \midrule
            $F1^{0.01}$ $\uparrow$ & 0.657 & 0.730 & 0.699 & 0.849 & \textbf{0.878} \\
            \midrule
            NC $\uparrow$ & 0.934 & 0.943 & 0.907 & 0.962 & \textbf{0.963} \\
            \bottomrule
    \end{tabular}}
  \caption{Evaluations on real scanned data. The evaluation metrics presented in the table represent the average results of these models.}
  \vspace{-0.2in}
  \label{fig:re-srb}
\end{figure}

\noindent\textbf{Real scanned data.} Dataset~\cite{10.1145/2451236.2451246} provide several real scanned point clouds, as illustrated in \cref{fig:re-srb}, we evaluate our model on the dataset to demonstrate the effectiveness. Our approach can reconstruct smooth surfaces from scanned data containing noise and outliers. However, our model cannot address the issue of missing parts. This limitation is due to the local geometric training strategy, which is independent of the global shape. 

\subsection{Analysis}
\textbf{Efficiency}. 
We compare the time complexity of our method with other methods, as shown in \cref{tab:time}. All tests were conducted on an Intel i9-13900K CPU and an NVIDIA RTX 4090 GPU. Computational results show that supervised, local feature-based methods like our approach and GeoUDF~\cite{ren2023geoudf} significantly outperform unsupervised methods in terms of computational efficiency. Additionally, our method has a significant improvement in training efficiency compared to GeoUDF. Utilizing ShapeNet as the training dataset, GeoUDF requires 120GB of storage space. In contrast, our method employs a shape-category-independent dataset, occupying merely 0.50GB of storage. Our network is very lightweight, with only 653KB of trainable parameters and a total parameter size of just 2MB.
Compared to GeoUDF, which requires 36 hours for training, our method only requires 14.5 hours. 

\noindent\textbf{Patch radius and point density.}
During the evaluation phase, the radius $r$ used to find the nearest points for each query point determines the size of the extracted patch and the range of effective query points in the space. The choice of radius directly influences the complexity of the geometric features captured. When normalizing point clouds to a unit bounding box, we set the radius, $r=0.018$. This setting achieves satisfactory reconstruction for our testing datasets. In the supplementary materials, we present a bias analysis experiment comparing the synthesized local patches and the local geometries extracted from the test point cloud data. The experimental results confirm that setting $r$ to 0.018 maintains a relatively low bias, suggesting its effectiveness. Users can conduct a preliminary bias analysis based on our well-trained model to adjust the size of the radius according to the complexity of the input point cloud. This process is not time-consuming.
Through experimental testing (refer to the supplementary materials), our algorithm ensures reasonable reconstruction, provided that there are at least 30 points within a unit area. A possible way for mitigating issues arising from low sampling rates is to apply an upsampling module~\cite{ren2023geoudf} during the pre-processing step.

\begin{figure}[t]
  \centering
  \includegraphics[width=0.9\linewidth]{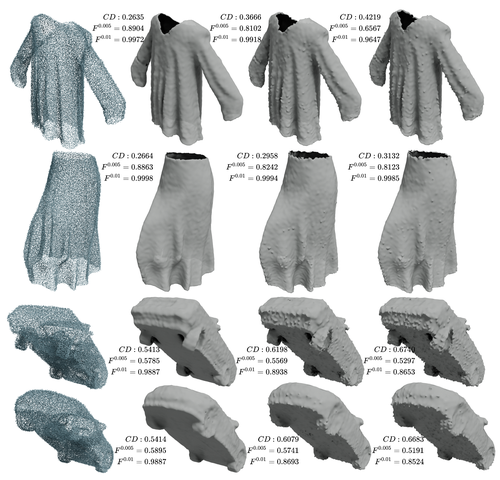}
  \\
  \vspace{-0.15in}
  \makebox[0.24\linewidth]{\footnotesize Input}
  \makebox[0.24\linewidth]{\footnotesize Ours}
  \makebox[0.24\linewidth]{\footnotesize Ablation (a)}
  \makebox[0.24\linewidth]{\footnotesize Ablation (b)}
  \vspace{-0.12in}
  \caption{Ablation studies on Cross-Attn module.  (a) Without the Points-Net and Cross-Attn modules.  (b) Without the Cross-Attn module. (CD score is multiplied by 100)}
  \vspace{-0.22in}
  \label{fig:net-abla}
\end{figure}
\subsection{Ablation studies}
\textbf{Cross-Attn module.}
Our main goal is to derive the UDF value for a query point by learning the local geometry within a radius $r$. To achieve this, we utilize Points-Net to capture the point cloud features $\mathbf{f}_p$ of local patches. This process enables the local geometry extracted from test data to align with the synthetic data through feature matching, even in the presence of noise or outliers. Vectors-Net is tasked with learning the features $\mathbf{f}_v$ of the set of vectors pointing towards the query point, which includes not only the position of the query point but also its distance information. The Cross-Attn module then processes these local patch features $\mathbf{f}_p$ as keys and values to query the vector features $\mathbf{f}_v$, which contain distance information, returning the most relevant feature $\mathbf{f}_G$ that determines the UDF value. See \cref{fig:net-abla} for two ablation studies on noisy point clouds.

\begin{figure}[htbp]
    \centering
    \includegraphics[width=0.9\linewidth]{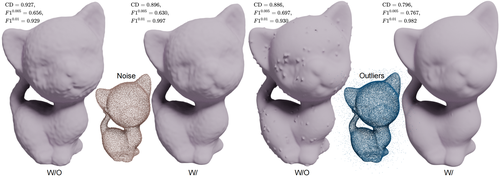}
    \vspace{-0.1in}
    \caption{Ablation on denoising module: Reconstructed surfaces from the same point clouds with noise/outliers corresponding to framework with and without the denoising module, respectively. }
    \vspace{-0.2in}
    \label{fig:abla-denoise}
\end{figure}

\noindent\textbf{Denoising module.} 
Our framework incorporates a denoising module to handle noisy point clouds. We conducted ablation experiments to verify the significance of this module. Specifically, we set $\lambda_d=0$ in the loss function to disable the denoising module, and then retrained the network. As illustrated in \cref{fig:abla-denoise}, we present the reconstructed surfaces for the same set of noisy point clouds with and without the denosing module, respectively. 

\subsection{Results of unsupervised integration}
\label{sec:integration}
\begin{figure}[!t]
    \centering
    \scriptsize
    \includegraphics[width=0.18\linewidth]{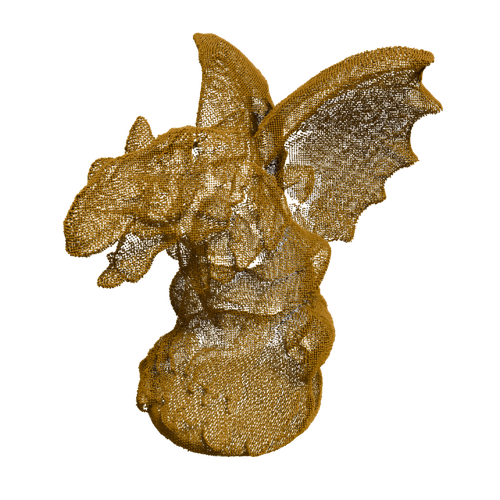}\hfill
    \includegraphics[width=0.18\linewidth]{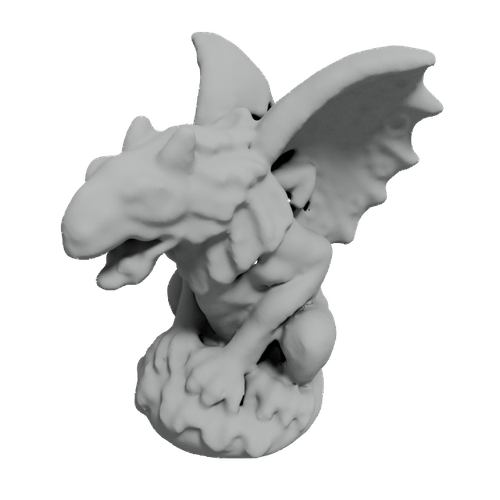}\hfill
    \includegraphics[width=0.18\linewidth]{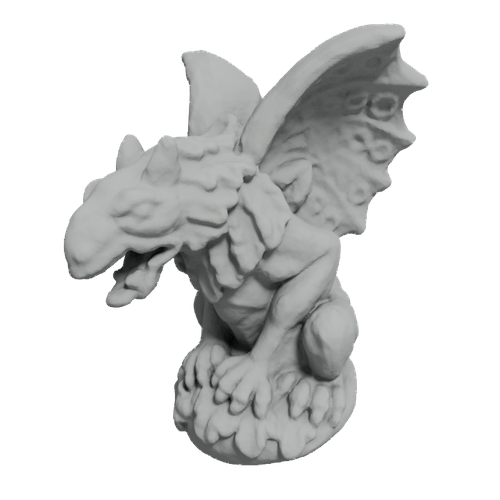}\hfill
    \includegraphics[width=0.18\linewidth]{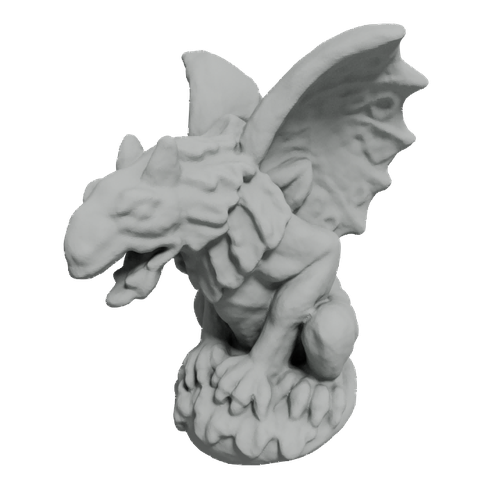}\hfill
    \includegraphics[width=0.18\linewidth]{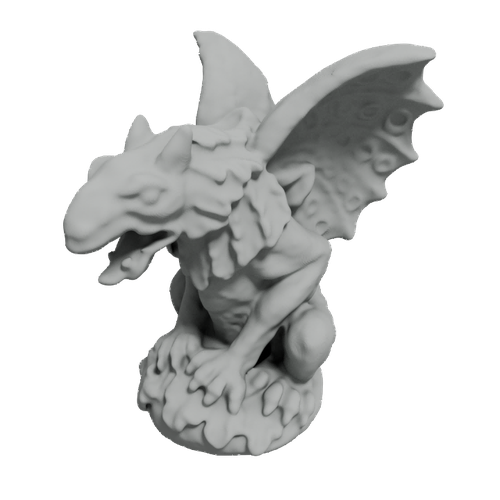}
    \vspace{-0.05in}
    \\
    \includegraphics[width=0.18\linewidth]{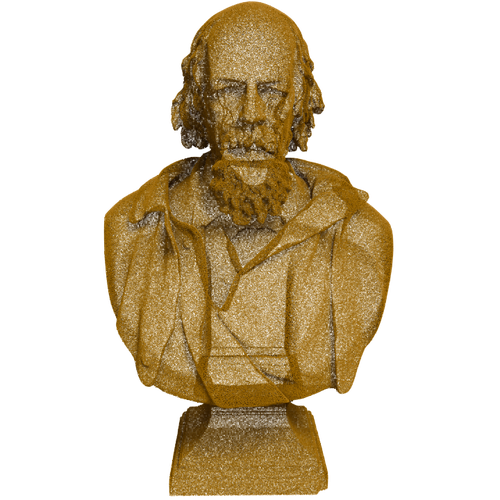}\hfill
    \includegraphics[width=0.18\linewidth]{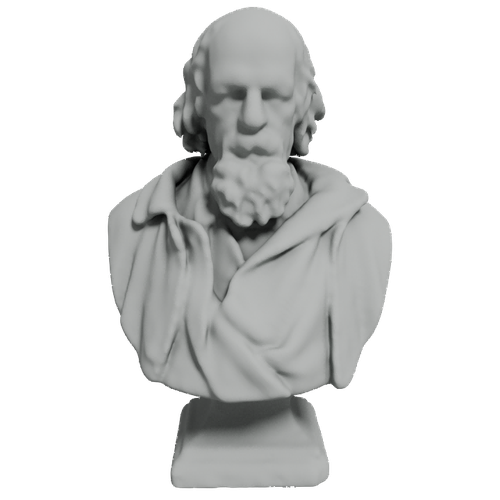}\hfill
    \includegraphics[width=0.18\linewidth]{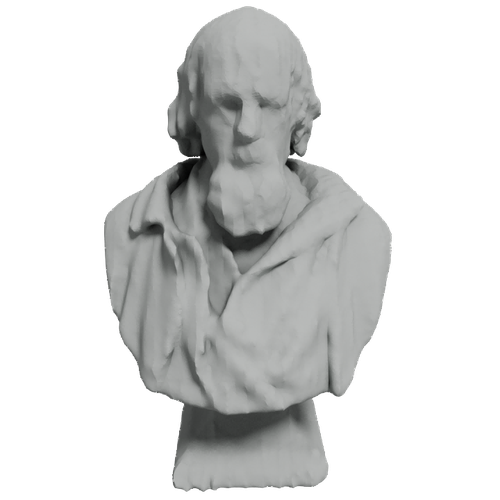}\hfill
    \includegraphics[width=0.18\linewidth]{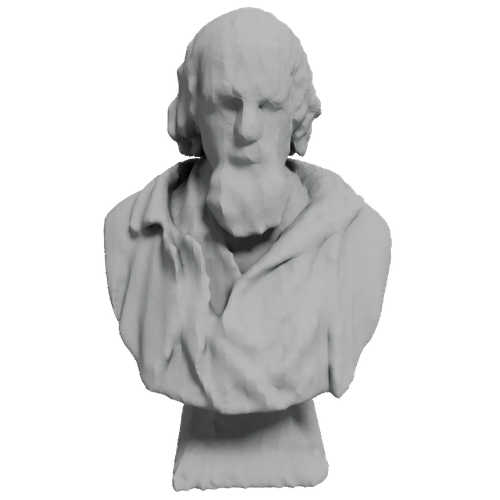}\hfill
    \includegraphics[width=0.18\linewidth]{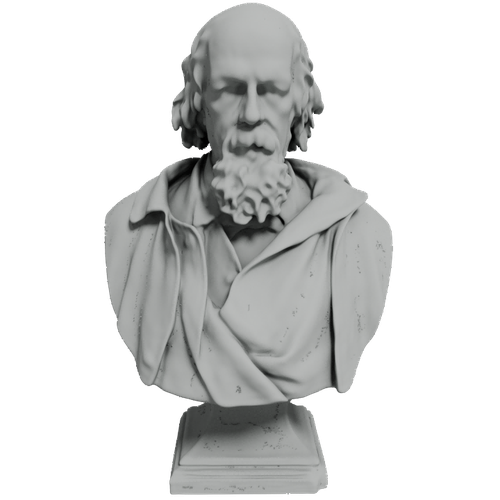}
    \vspace{-0.04in}
    \\
    \includegraphics[width=0.18\linewidth]{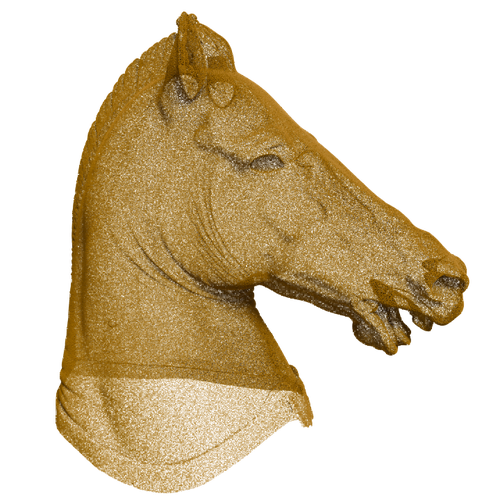}\hfill
    \includegraphics[width=0.18\linewidth]{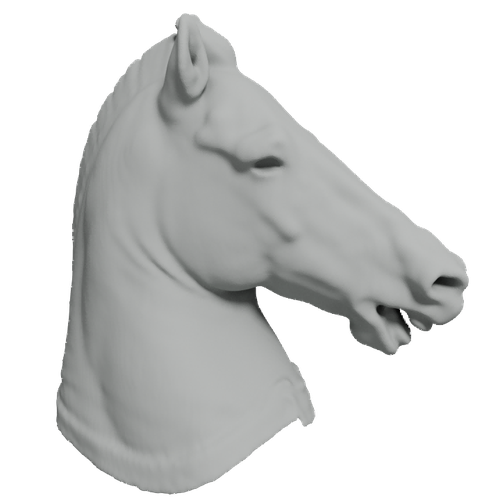}\hfill
    \includegraphics[width=0.18\linewidth]{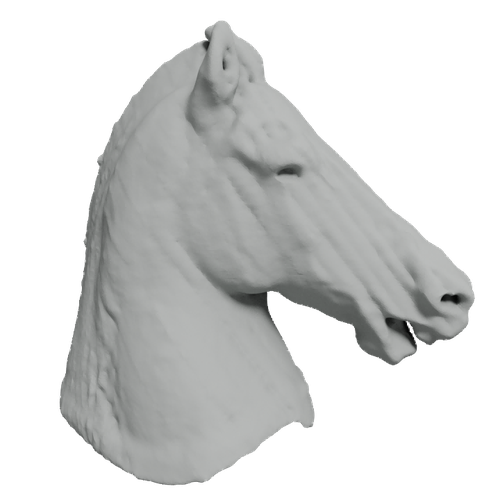}\hfill
    \includegraphics[width=0.18\linewidth]{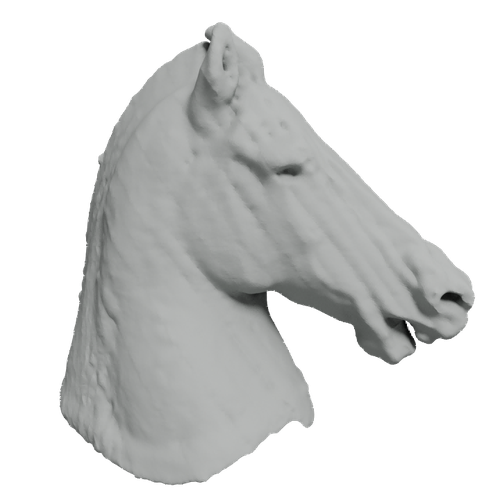}\hfill
    \includegraphics[width=0.18\linewidth]{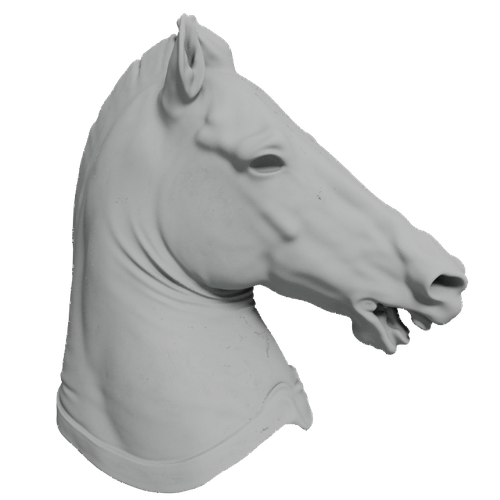}
    \vspace{-0.06in}
    \\
    \includegraphics[width=0.18\linewidth]{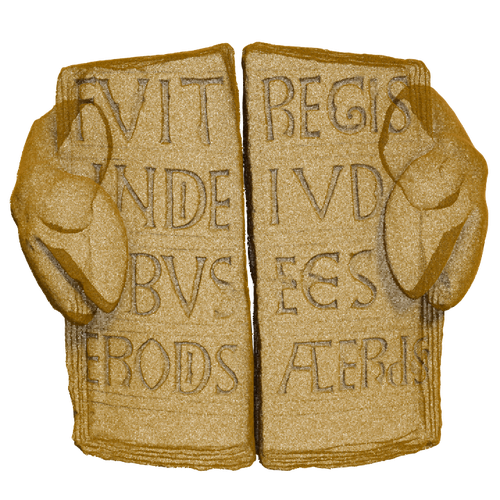}\hfill
    \includegraphics[width=0.18\linewidth]{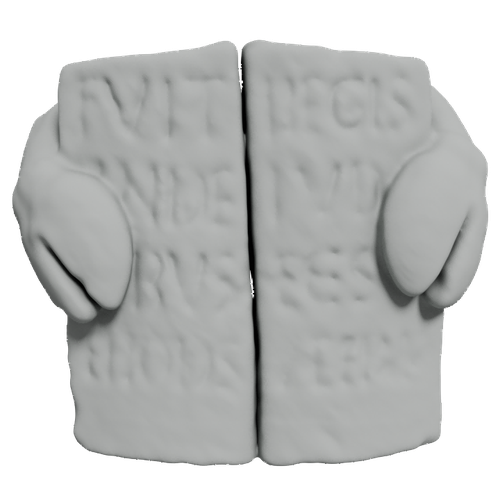}\hfill
    \includegraphics[width=0.18\linewidth]{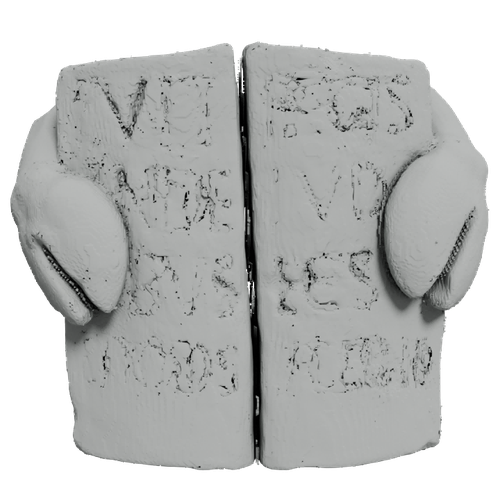}\hfill
    \includegraphics[width=0.18\linewidth]{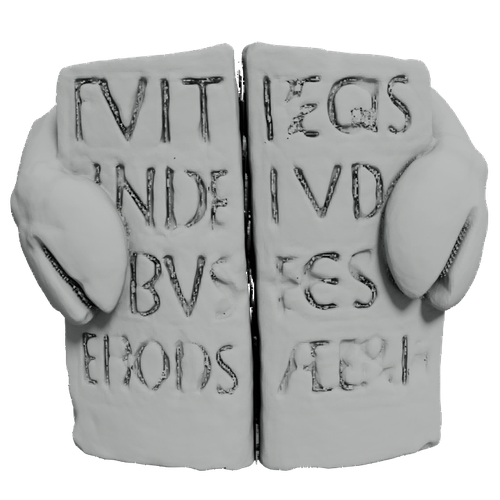}\hfill
    \includegraphics[width=0.18\linewidth]{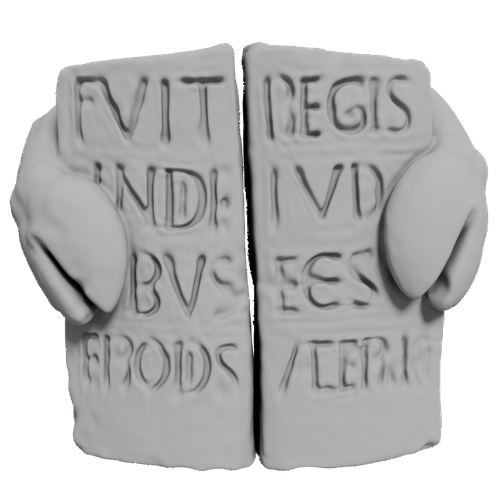}
    \vspace{-0.06in}
    \\
    \includegraphics[width=0.18\linewidth]{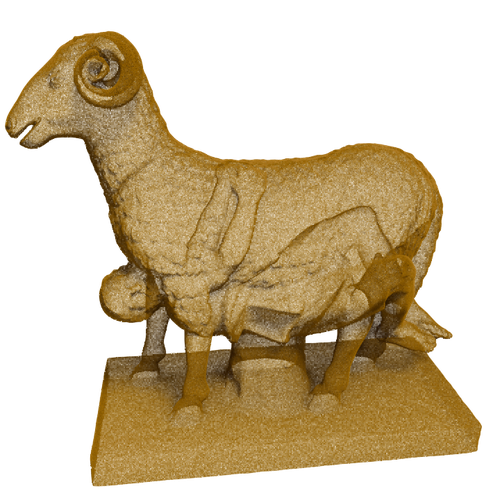}\hfill
    \includegraphics[width=0.18\linewidth]{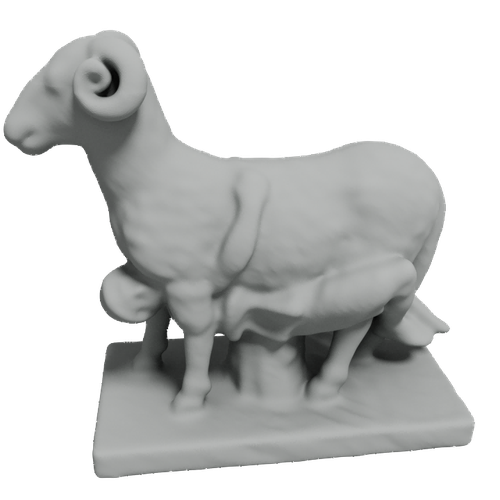}\hfill
    \includegraphics[width=0.18\linewidth]{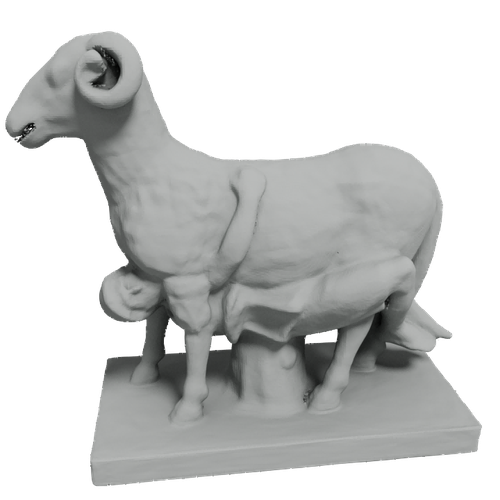}\hfill
    \includegraphics[width=0.18\linewidth]{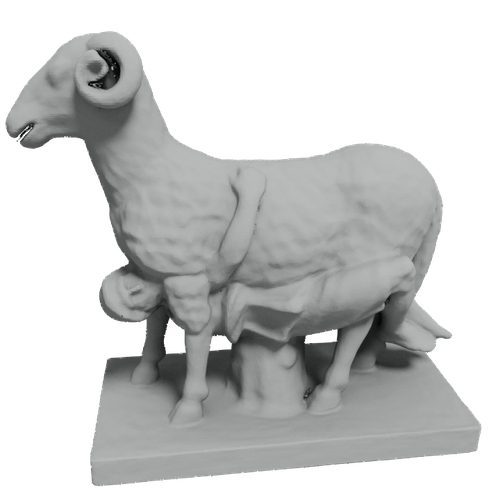}\hfill
    \includegraphics[width=0.18\linewidth]{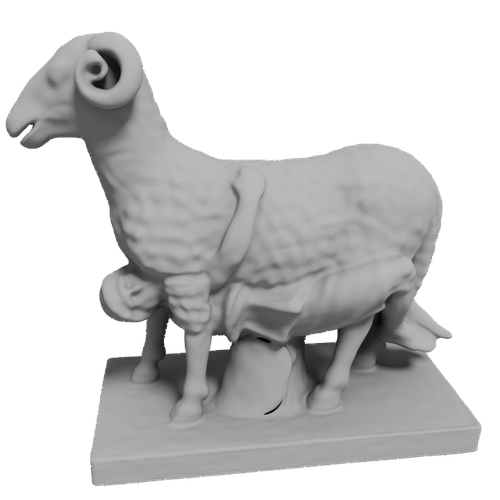}
    \vspace{-0.06in}
    \\
    \makebox[0.18\linewidth]{Input}\hfill
    \makebox[0.18\linewidth]{LoSF}\hfill
    \makebox[0.18\linewidth]{*+CAP-UDF}\hfill
    \makebox[0.18\linewidth]{*+LevelSetUDF}\hfill
    \makebox[0.18\linewidth]{*+SIREN}
    \vspace{-0.05in}
    \caption{Reconstruction results of the integrated framework.}
    \vspace{-0.22in}
    \label{fig:integration}
\end{figure}

Our LoSF-UDF approach offers better initialization for unsupervised methods, including CAP-UDF~\cite{Zhou2022CAP-UDF}, LevelSetUDF~\cite{zhou2023levelset}, and DEUDF~\cite{deng20242sudf}. We evaluated 12 models on the Threescan dataset~\cite{threedscan}, each containing rich details. Using the integration framework based on our lightweight LoSF-UDF, we achieve comparable or even superior reconstruction results, as illustrated in \cref{fig:integration} and \cref{tab:inte-quanti}. More importantly, we improve the efficiency of original unsupervised methods as shown in \cref{tab:time}.
Considering the DEUDF is not open-source, we employ their proposed loss functions to train a SIREN network~\cite{siren} on our own. For the loss function terms that require normal information, we used the method of PCA~\cite{10.1145/133994.134011} to estimate the normals during the optimization process. Thanks to the robustness of LoSF, the accuracy of its estimation has been enhanced.
\begin{table}[!t]
  \centering
  \caption{Quantitative comparison results of the integrated framework with unsupervised methods.}
   \vspace{-0.1in}
  \resizebox{1.0\linewidth}{!}{\begin{tabular}{r c c c c c c}
            \toprule
            Method & LoSF & *+CAP-UDF~\cite{Zhou2022CAP-UDF} & *+LevelSetUDF~\cite{zhou2023levelset} & *+SIREN~\cite{xu2024detail}  \\
            \midrule
            CD($\times$100) $\downarrow$ & 0.409 & 0.447 & 0.429 & \textbf{0.217}  \\
            \midrule
            $F1^{0.005}$ $\uparrow$ & 0.638 & 0.609 & 0.610 & \textbf{0.906}  \\
            \midrule
            $F1^{0.01}$ $\uparrow$ & 0.958 & 0.936 & 0.952 & \textbf{0.984}  \\
            \midrule
            NC $\uparrow$ & 0.964 & 0.946 & 0.962 & \textbf{0.969}  \\
            \bottomrule
  \end{tabular}}
    \label{tab:inte-quanti}
    \vspace{-0.09in}
\end{table}

\section{Conclusion}
In this paper, we introduce a novel and lightweight neural framework for surface reconstruction from 3D point clouds by learning UDFs from local shape functions. Our key insight is that 3D shapes exhibit simple patterns within localized regions, which can be exploited to create a training dataset of point cloud patches represented by mathematical functions. As a result, our method enables efficient and robust surfaces reconstruction without the need for shape-specific training, even in the presence of noise and outliers. Extensive experiments on various datasets have demonstrated the efficacy of our method. Moreover, our lightweight framework can be integrated with unsupervised methods to provide rapid and reliable initialization, enhancing both efficiency and accuracy.

\clearpage
\newpage
\section*{Acknowledgement}
The NTU authors were supported in part by the Ministry of Education, Singapore, under its Academic Research Fund Grants (MOE-T2EP20220-0005 \& RT19/22) and the RIE2020 Industry Alignment Fund–Industry Collaboration Projects (IAF-ICP) Funding Initiative, as well as cash and in-kind contribution from the industry partner(s). The DLUT authors were supported by the National Natural Science Foundation of China under Grants 62402083 and T2225012, the National Key
R\&D Program of China under Grants 2021YFA1003003. F. Hou was supported by the Basic Research Project of ISCAS (ISCAS-JCMS-202303) and the Major Research Project of ISCAS (ISCAS-ZD-202401). J. Hou was supported by the Hong Kong RGC under Grants 11219422 and 11219324. 
{
    \small
    \bibliographystyle{ieeenat_fullname}
    \bibliography{main}
}

\end{document}


\clearpage
\setcounter{page}{1}
\maketitlesupplementary

\section{Bias analysis}
\label{sec:bias}
To illustrate the effectiveness of generalizing from our synthetic dataset to general point clouds, we conduct a bias analysis comparing our synthetic local patches with local geometries extracted from test data. We obtain the feature vectors $\mathbf{f}^i_p$ for all patches in the training dataset (totaling 131,064 data points), processed by Points-Net in our network, and then extract local geometries with a radius $r=0.018$ from point clouds in three categories: DeepFashion3D, ShapeNet-Cars, and 10 common computer graphics models (Bunny, Bimba, etc). 
 We compute the corresponding feature vectors $\hat{\mathbf{f}}^i_p$, and measure the distances between $\mathbf{f}^i_p$ and $\hat{\mathbf{f}}^i_p$ as the bias. The same analysis is also applied to point clouds with noise and outliers. As illustrated in \cref{fig:bias} (a), we employ boxplots to visualize the bias distribution. The results indicate that bias remains low under conditions that are clean, noisy, or affected by outliers. The radius impacts the size of extracted local geometries and thus influences the observed bias. We performed experiments with different radii using the aforementioned analysis method. \cref{fig:bias} (b) shows a radius of $r=0.018$ (our default setting) results in relatively fewer outliers.
\begin{figure}[htbp]
    \centering
    \includegraphics[width=\linewidth]{supp_files/data-boxplot.png}\\
    \makebox[\linewidth]{(a) Bias analysis for point clouds from three datasets}
    \includegraphics[width=\linewidth]{supp_files/radiux-boxplot.png}
    \\
    \makebox[\linewidth]{(b) Joint analysis with radius}
    \caption{We statistically analyze the feature bias between local geometries extracted from test data and local patches from our synthetic training dataset, illustrating the bias distributions with box plots. (a) The bias remains low across three different datasets.(b) A radius of 0.018 results in relatively fewer outliers in most scenarios.}
    \label{fig:bias}
\end{figure}

\section{Radius and point cloud density}
Given that our method learns UDF values from locally extracted patches in input point clouds, the density and radius of these clouds can affect the reconstruction results. To evaluate the applicability of our method's parameters, we conduct performance evaluation experiments under varying point cloud densities and radii. \Cref{fig:bunny-density} presents several examples from our experiments. We observed that the maximum applicable radius is $r=0.05$. This means that when $r$ exceeds 0.05, the feature information within the local geometry surpasses our network's ability to evaluate, preventing accurate reconstruction of the geometry, as illustrated in the upper right area of \cref{fig:bunny-density}. In order to standardize the point cloud density scale, we count the number of points within a circle of radius 0.05 as the density value. Experiments demonstrate that our method cannot successfully reconstruct when the density is approximately less than 30 , as shown in the lower left area of \cref{fig:bunny-density}.
Therefore, for user-inputted point clouds, the density of the point cloud is first assessed. If it is below the algorithm's requirements, upsampling can be used to densify the point cloud. For general models, a radius setting of 0.018 is feasible in most cases. Users can also quickly perform bias analysis and make appropriate adjustments within a range of less than 0.05.

\begin{figure*}
    \centering
    \includegraphics[width=0.33\textwidth]{supp_files/bunny-radius-density.png}\hfill
    \includegraphics[width=0.33\textwidth]{supp_files/cloth-radius-density.png}\hfill
    \includegraphics[width=0.33\textwidth]{supp_files/cars-radius-density.png}
    \\
    \makebox[0.33\textwidth]{(a) Bunny}\hfill
    \makebox[0.33\textwidth]{(b) DeepFashion}\hfill
    \makebox[0.33\textwidth]{(c) ShapeNetCars}
    \caption{We evaluate the performance of our algorithm across varying point cloud densities and radii.}
    \label{fig:bunny-density}
\end{figure*}

\begin{figure*}[htbp]
    \centering
    \includegraphics[width=1.0\linewidth]{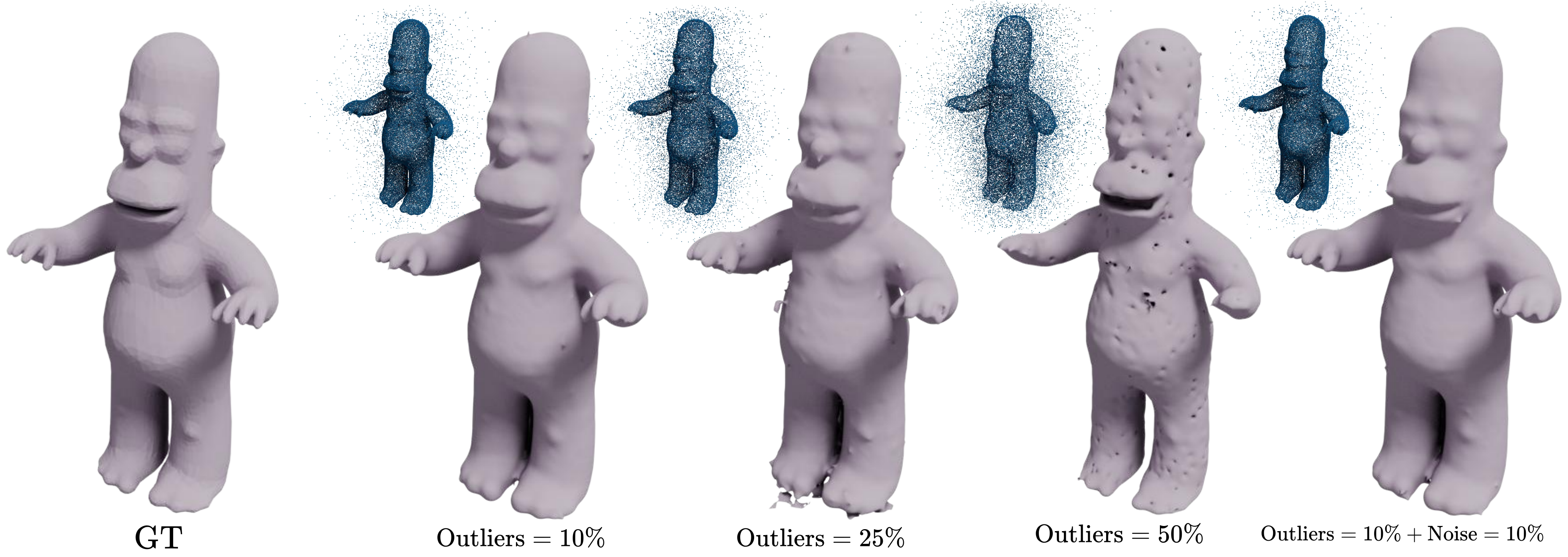}
    \caption{Our model demonstrates robustness to more outliers.}
    \label{fig:more-outliers}
\end{figure*}
\begin{figure*}
    \centering
    \includegraphics[width=1\textwidth]{supp_files/indoor_scene_losf_reconstruction.png}
    \caption{Reconstruction result on large scene dataset}
    \label{fig:scene-reconstrustion}
\end{figure*}

\section{More results}
\begin{figure*}[h]
    \centering
    \includegraphics[width=0.137\linewidth]{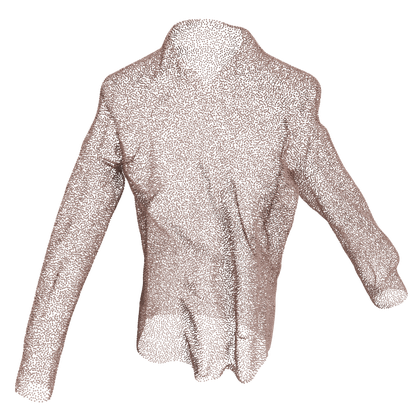}
    \hfill
    \includegraphics[width=0.137\linewidth]{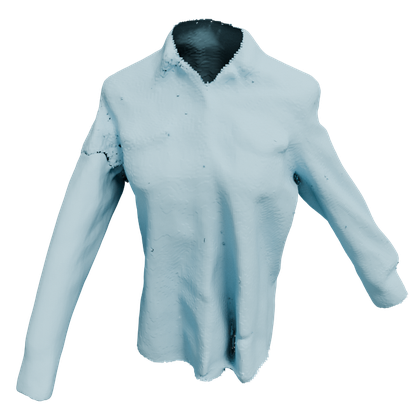}
    \hfill
    \includegraphics[width=0.137\linewidth]{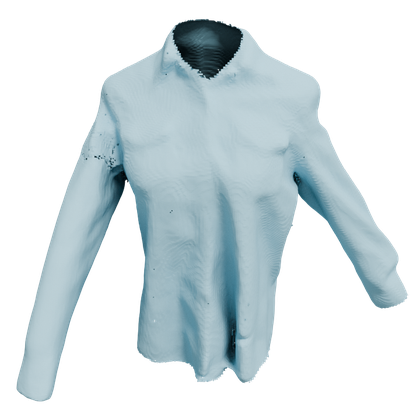}
    \hfill
    \includegraphics[width=0.137\linewidth]{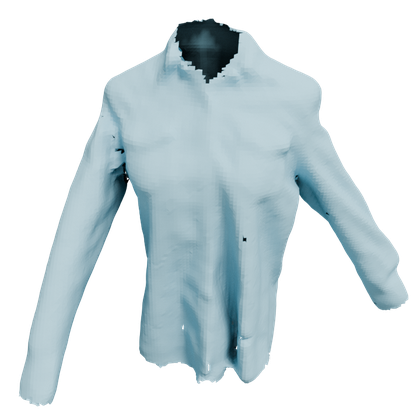}
    \hfill
    \includegraphics[width=0.137\linewidth]{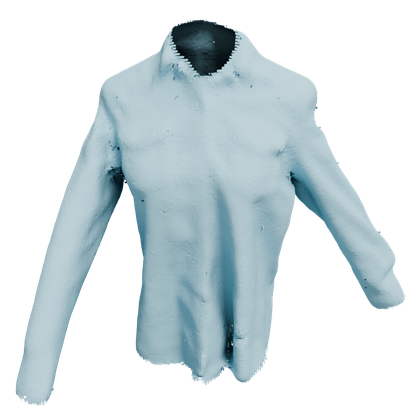}
    \hfill
    \includegraphics[width=0.137\linewidth]{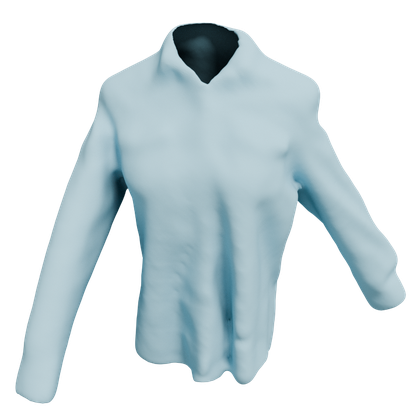}
    \hfill
    \includegraphics[width=0.137\linewidth]{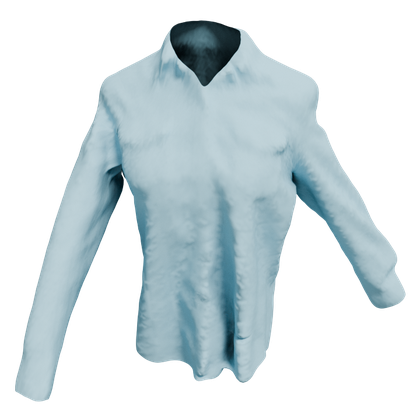}
    \\
    
    \includegraphics[width=0.137\linewidth]{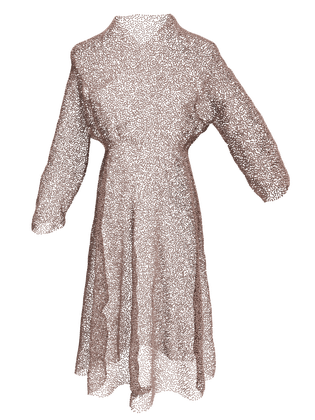}
    \hfill
    \includegraphics[width=0.137\linewidth]{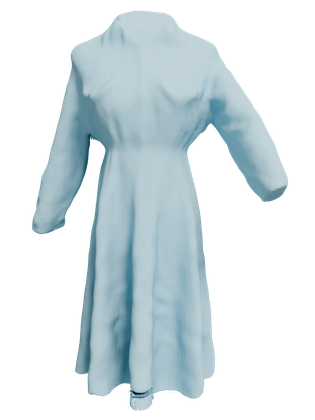}
    \hfill
    \includegraphics[width=0.137\linewidth]{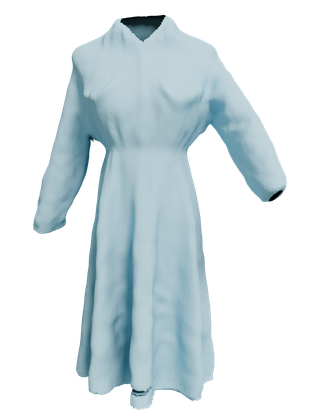}
    \hfill
    \includegraphics[width=0.137\linewidth]{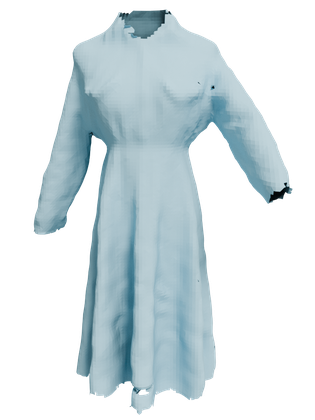}
    \hfill
    \includegraphics[width=0.137\linewidth]{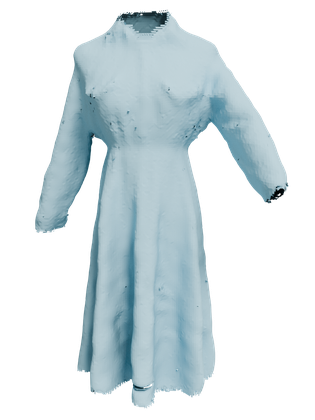}
    \hfill
    \includegraphics[width=0.137\linewidth]{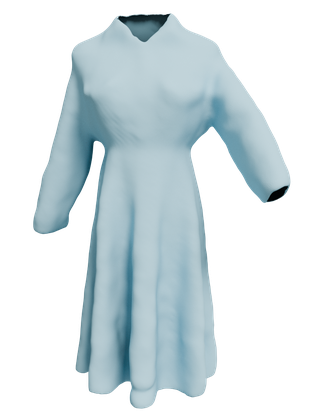}
    \hfill
    \includegraphics[width=0.137\linewidth]{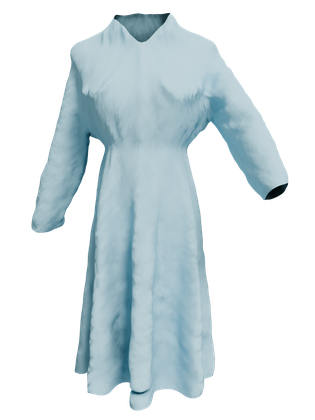}
    \\
    
    \includegraphics[width=0.137\linewidth]{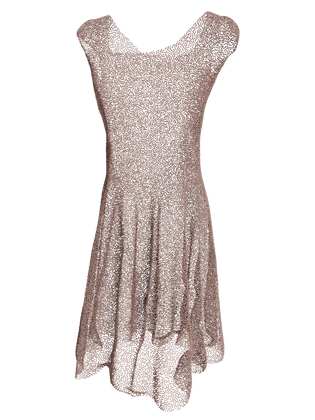}
    \hfill
    \includegraphics[width=0.137\linewidth]{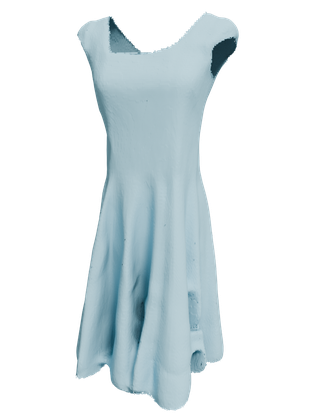}
    \hfill
    \includegraphics[width=0.137\linewidth]{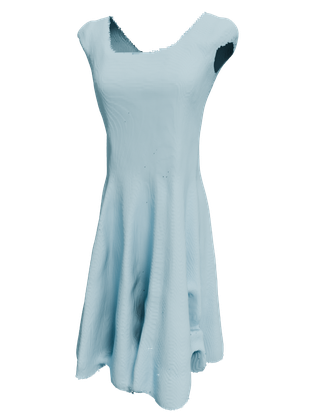}
    \hfill
    \includegraphics[width=0.137\linewidth]{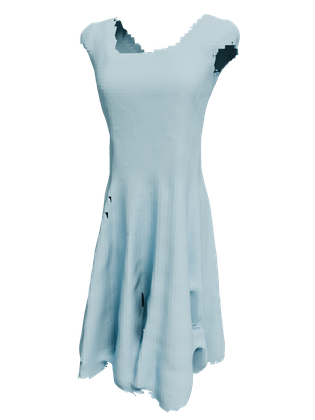}
    \hfill
    \includegraphics[width=0.137\linewidth]{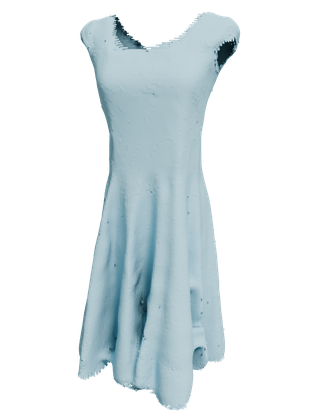}
    \hfill
    \includegraphics[width=0.137\linewidth]{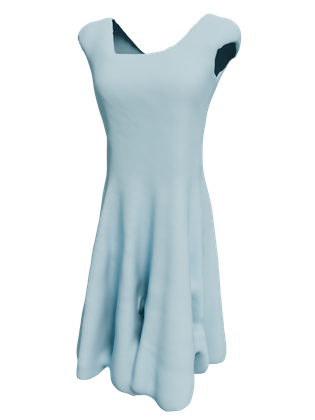}
    \hfill
    \includegraphics[width=0.137\linewidth]{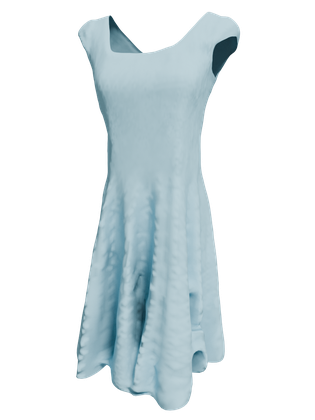}
    \\
    
    \includegraphics[width=0.137\linewidth]{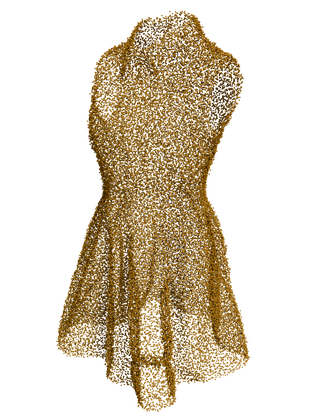}
    \hfill
    \includegraphics[width=0.137\linewidth]{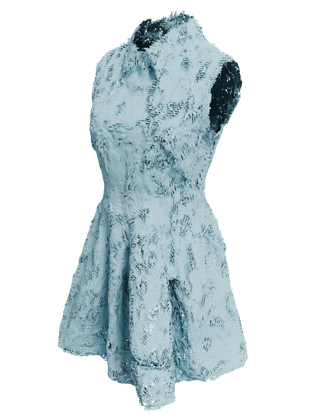}
    \hfill
    \includegraphics[width=0.137\linewidth]{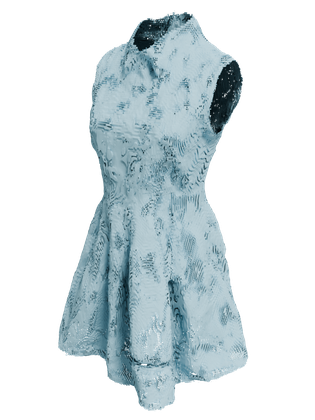}
    \hfill
    \includegraphics[width=0.137\linewidth]{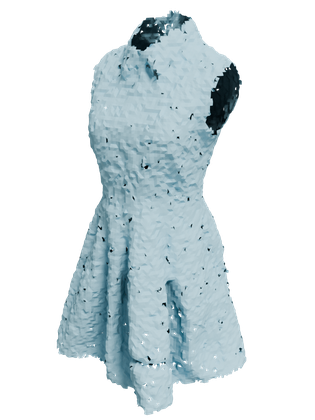}
    \hfill
    \includegraphics[width=0.137\linewidth]{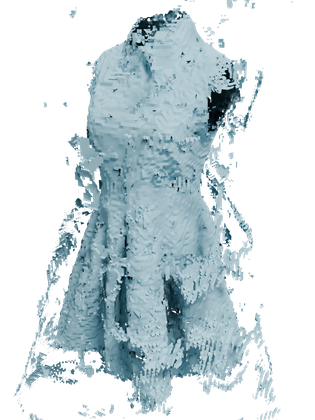}
    \hfill
    \includegraphics[width=0.137\linewidth]{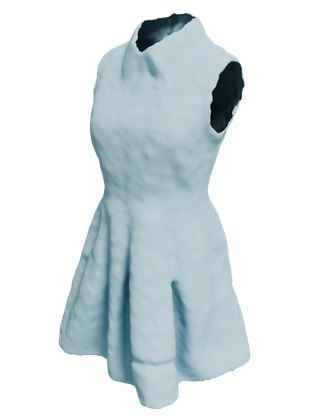}
    \hfill
    \includegraphics[width=0.137\linewidth]{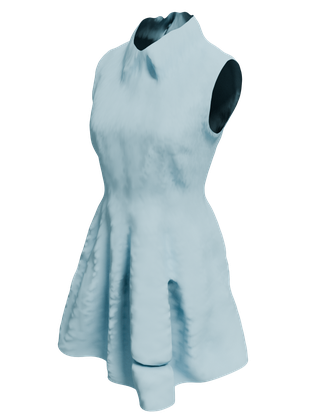}
    \\
    
    \includegraphics[width=0.137\linewidth]{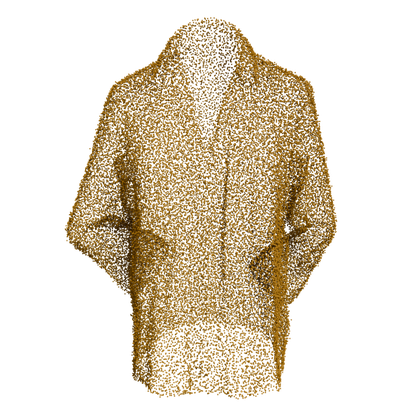}
    \hfill
    \includegraphics[width=0.137\linewidth]{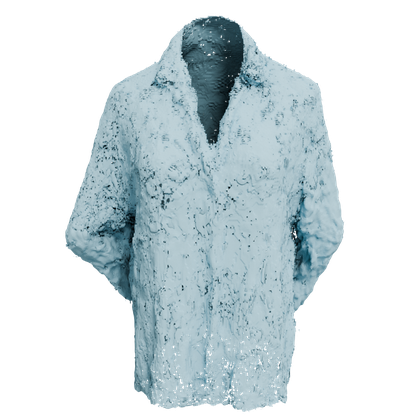}
    \hfill
    \includegraphics[width=0.137\linewidth]{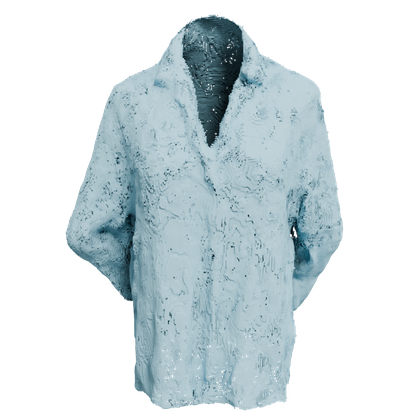}
    \hfill
    \includegraphics[width=0.137\linewidth]{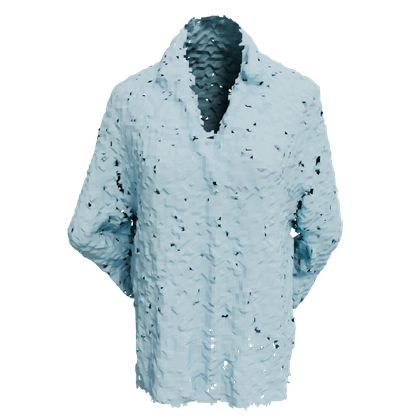}
    \hfill
    \includegraphics[width=0.137\linewidth]{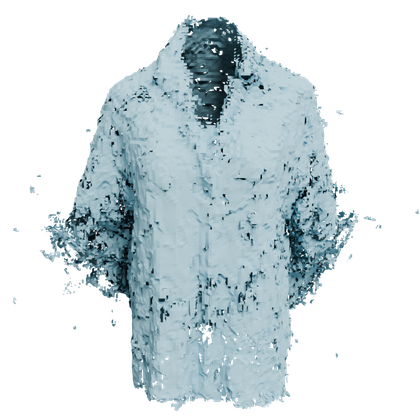}
    \hfill
    \includegraphics[width=0.137\linewidth]{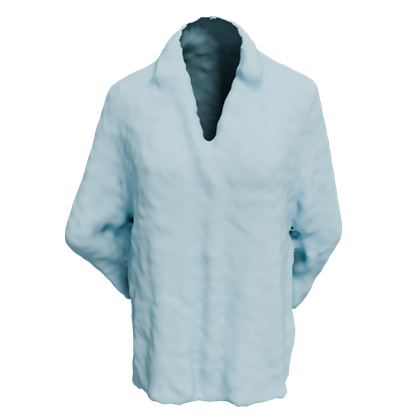}
    \hfill
    \includegraphics[width=0.137\linewidth]{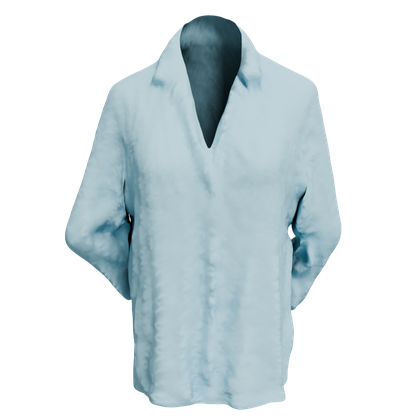}
    \\
    
    \includegraphics[width=0.137\linewidth]{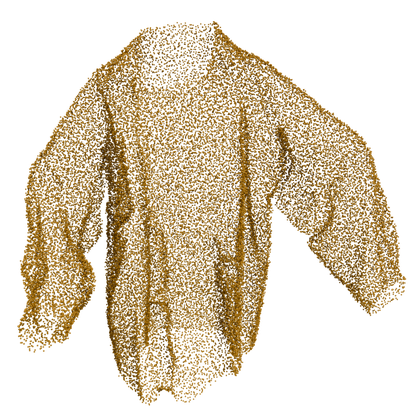}
    \hfill
    \includegraphics[width=0.137\linewidth]{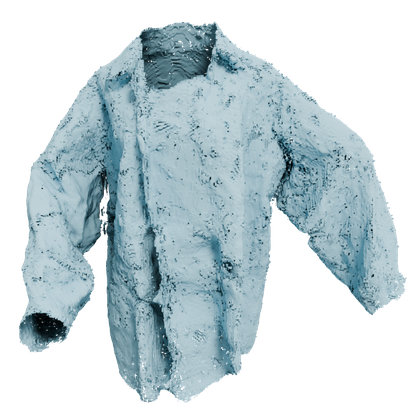}
    \hfill
    \includegraphics[width=0.137\linewidth]{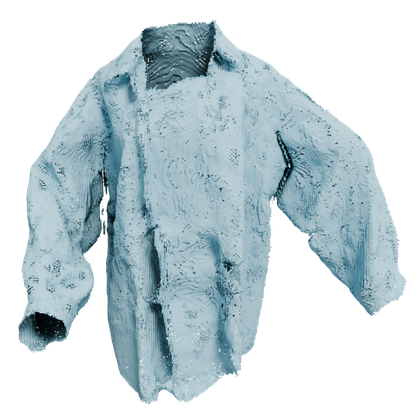}
    \hfill
    \includegraphics[width=0.137\linewidth]{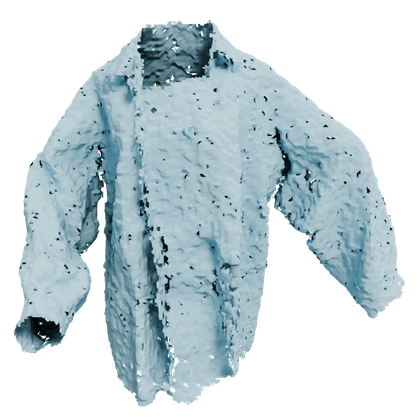}
    \hfill
    \includegraphics[width=0.137\linewidth]{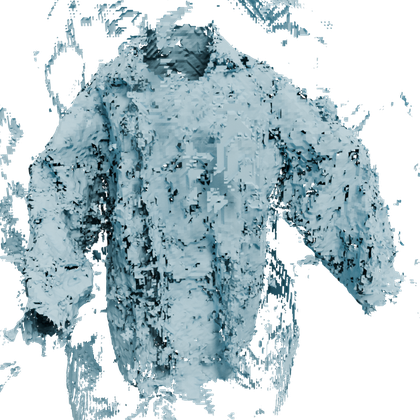}
    \hfill
    \includegraphics[width=0.137\linewidth]{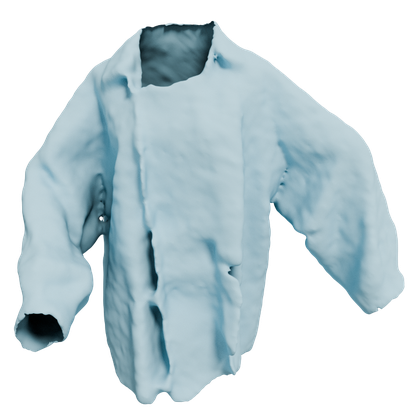}
    \hfill
    \includegraphics[width=0.137\linewidth]{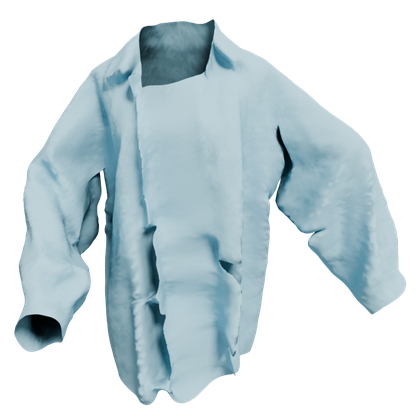}
    \\

    \makebox[0.137\linewidth]{\small(a) \small{Input}}
    \hfill
    \makebox[0.137\linewidth]{\scriptsize (b) \scriptsize {CAP-UDF}}
    \hfill
    \makebox[0.137\linewidth]{\scriptsize (c) \scriptsize{LevelSetUDF}}
    \hfill
    \makebox[0.137\linewidth]{\small(d) \small{GeoUDF}}
    \hfill
    \makebox[0.137\linewidth]{\small(e) \small{DUDF}}
    \hfill
    \makebox[0.137\linewidth]{\small(f) \small Ours}
    \hfill
    \makebox[0.137\linewidth]{\small(g) \small{GT}}
    \caption{More visual results on the DeepFashion3D dataset. Top three rows: Reconstruction results under noise-free conditions. Bottom three rows: Reconstruction results under noise condition.}
    \label{fig:more-clean}
\end{figure*}

\begin{figure*}[h]
    \centering
    \includegraphics[width=0.137\linewidth]{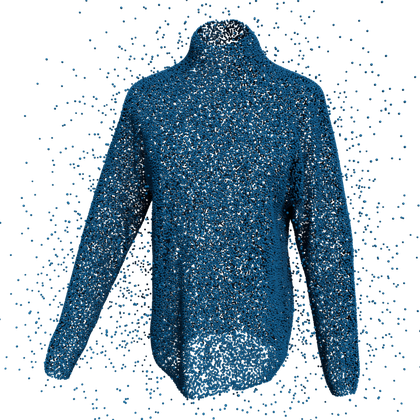}
    \hfill
    \includegraphics[width=0.137\linewidth]{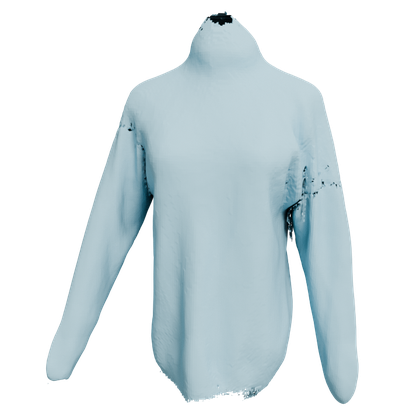}
    \hfill
    \includegraphics[width=0.137\linewidth]{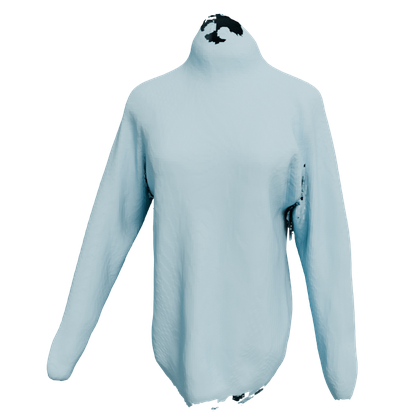}
    \hfill
    \includegraphics[width=0.137\linewidth]{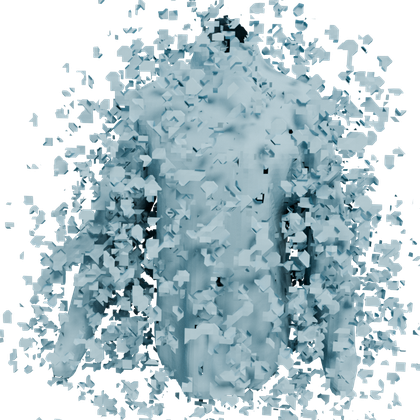}
    \hfill
    \includegraphics[width=0.137\linewidth]{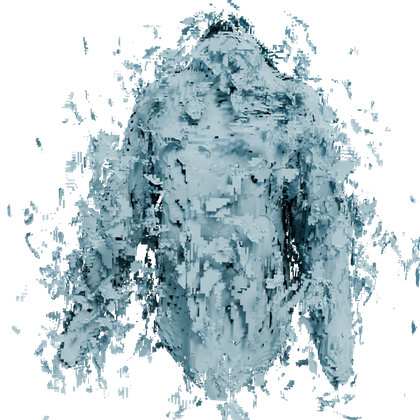}
    \hfill
    \includegraphics[width=0.137\linewidth]{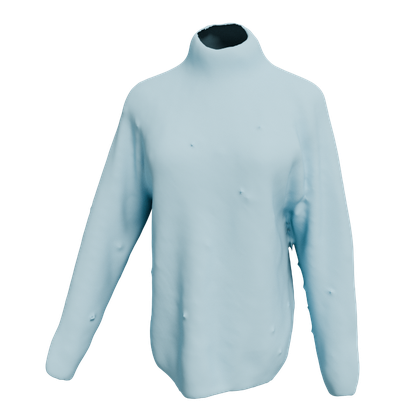}
    \hfill
    \includegraphics[width=0.137\linewidth]{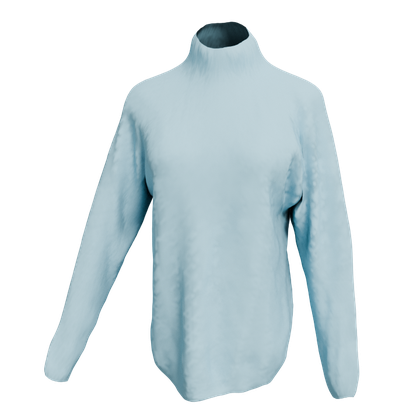}
    \\
    
    \includegraphics[width=0.137\linewidth]{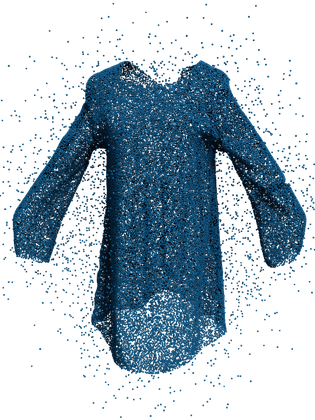}
    \hfill
    \includegraphics[width=0.137\linewidth]{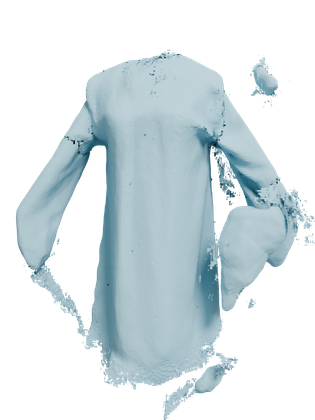}
    \hfill
    \includegraphics[width=0.137\linewidth]{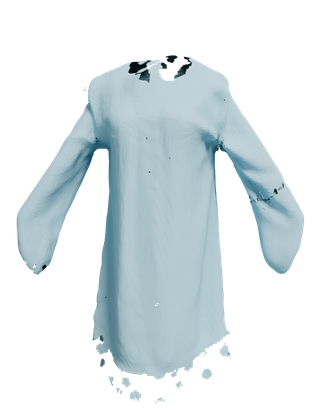}
    \hfill
    \includegraphics[width=0.137\linewidth]{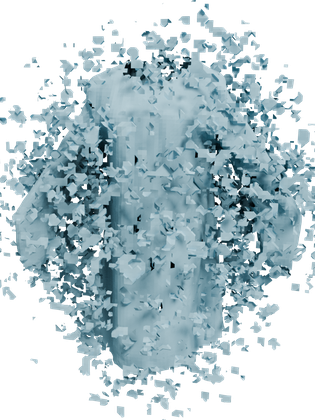}
    \hfill
    \includegraphics[width=0.137\linewidth]{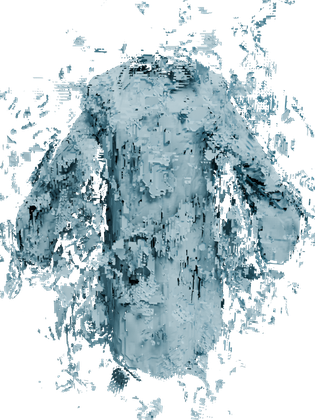}
    \hfill
    \includegraphics[width=0.137\linewidth]{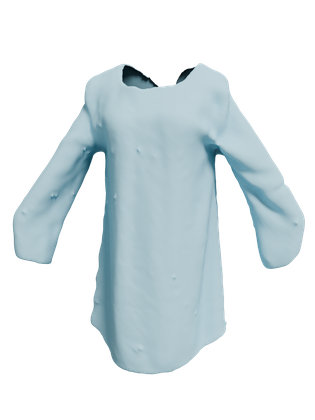}
    \hfill
    \includegraphics[width=0.137\linewidth]{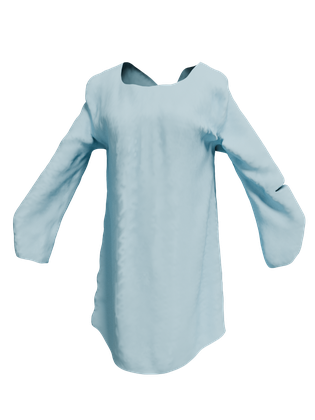}
    \\
    
    \includegraphics[width=0.137\linewidth]{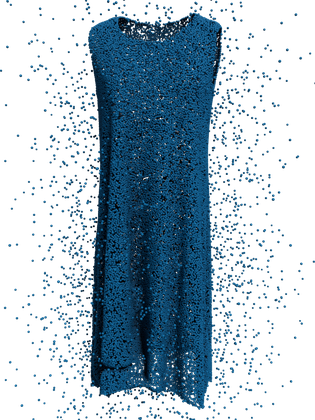}
    \hfill
    \includegraphics[width=0.137\linewidth]{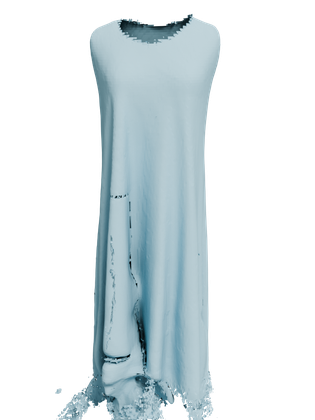}
    \hfill
    \includegraphics[width=0.137\linewidth]{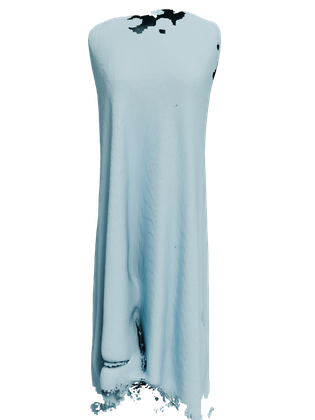}
    \hfill
    \includegraphics[width=0.137\linewidth]{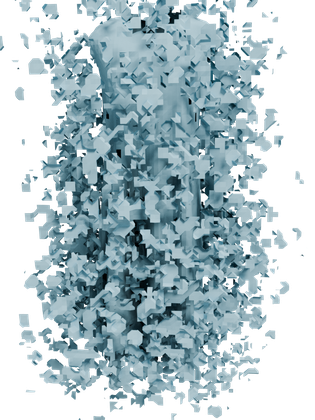}
    \hfill
    \includegraphics[width=0.137\linewidth]{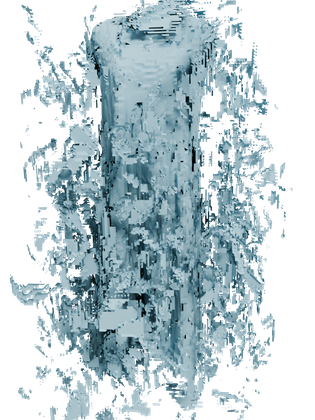}
    \hfill
    \includegraphics[width=0.137\linewidth]{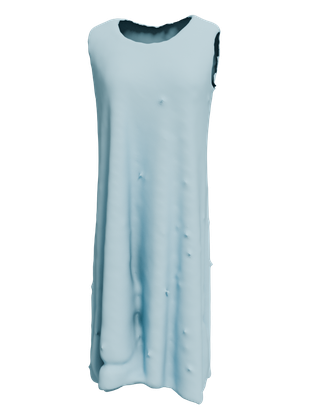}
    \hfill
    \includegraphics[width=0.137\linewidth]{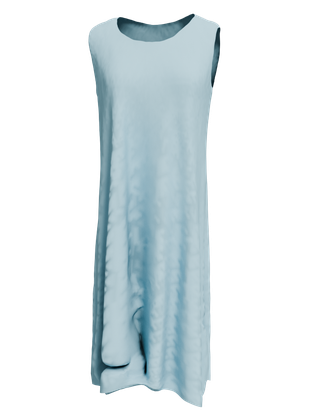}
    \\
    
    \includegraphics[width=0.137\linewidth]{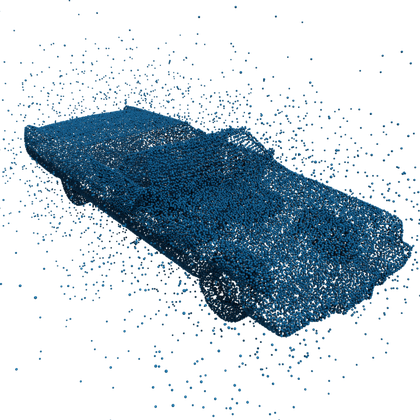}
    \hfill
    \includegraphics[width=0.137\linewidth]{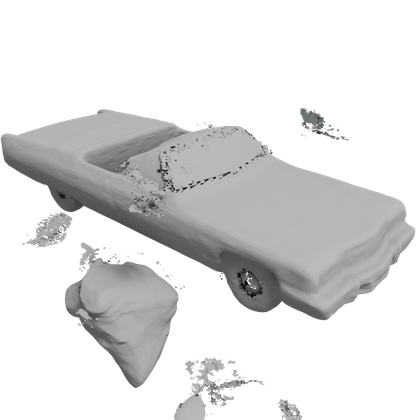}
    \hfill
    \includegraphics[width=0.137\linewidth]{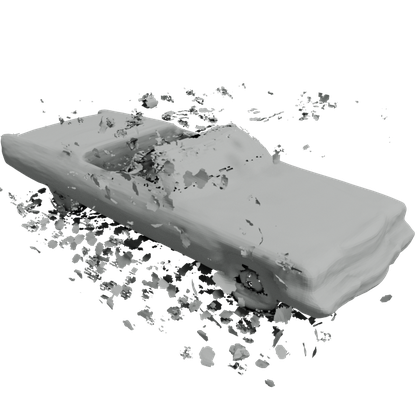}
    \hfill
    \includegraphics[width=0.137\linewidth]{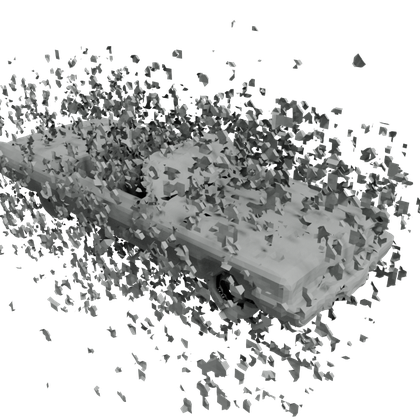}
    \hfill
    \includegraphics[width=0.137\linewidth]{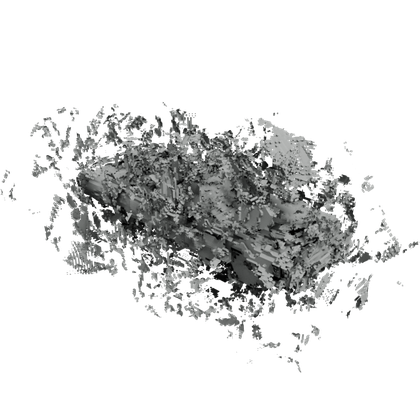}
    \hfill
    \includegraphics[width=0.137\linewidth]{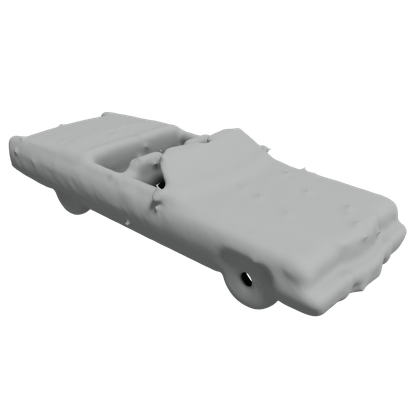}
    \hfill
    \includegraphics[width=0.137\linewidth]{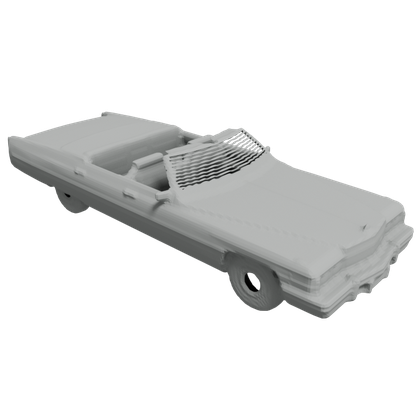}
    \\
    
    \includegraphics[width=0.137\linewidth]{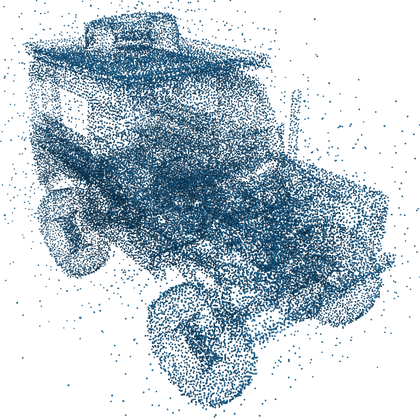}
    \hfill
    \includegraphics[width=0.137\linewidth]{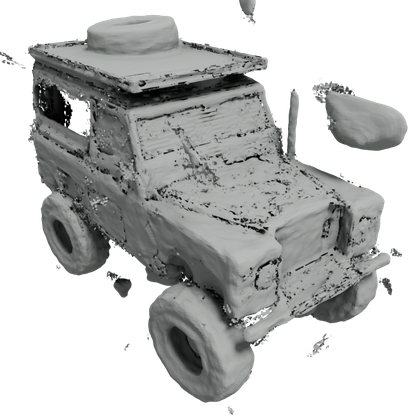}
    \hfill
    \includegraphics[width=0.137\linewidth]{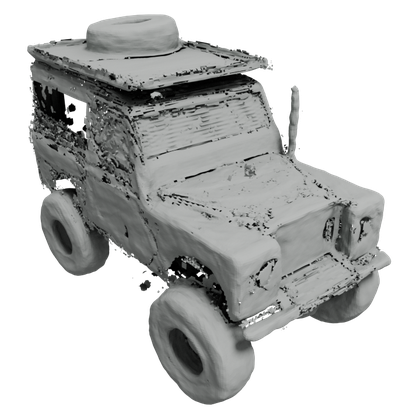}
    \hfill
    \includegraphics[width=0.137\linewidth]{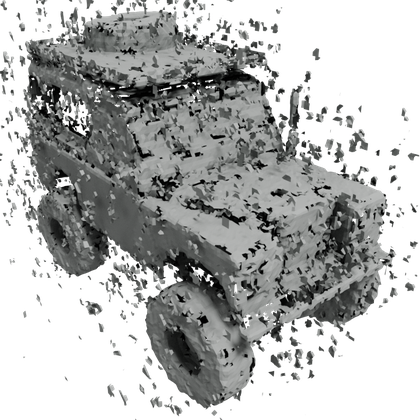}
    \hfill
    \includegraphics[width=0.137\linewidth]{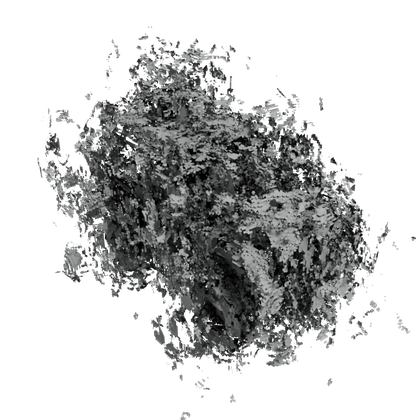}
    \hfill
    \includegraphics[width=0.137\linewidth]{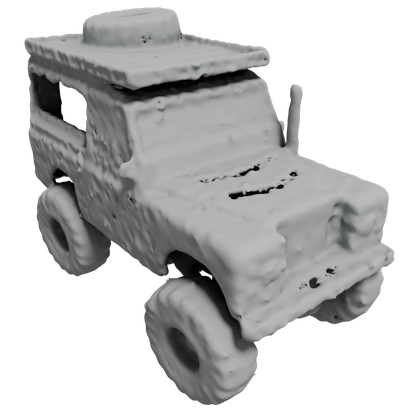}
    \hfill
    \includegraphics[width=0.137\linewidth]{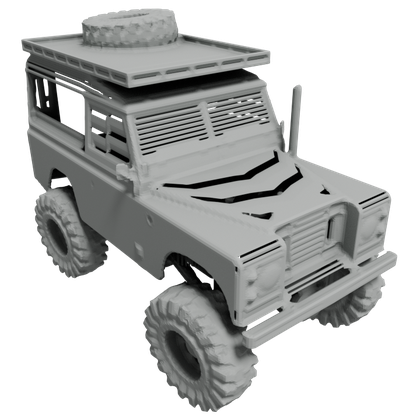}
    \\
    
    \includegraphics[width=0.137\linewidth]{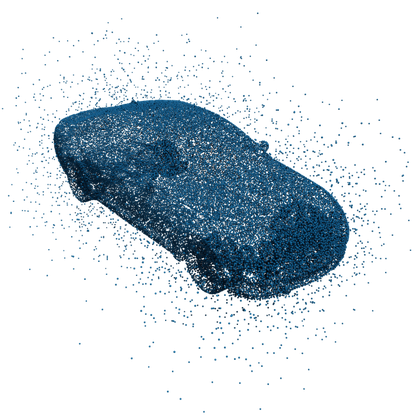}
    \hfill
    \includegraphics[width=0.137\linewidth]{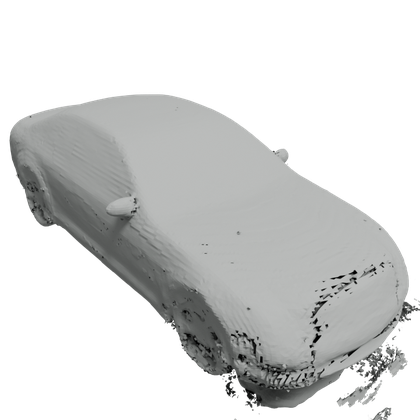}
    \hfill
    \includegraphics[width=0.137\linewidth]{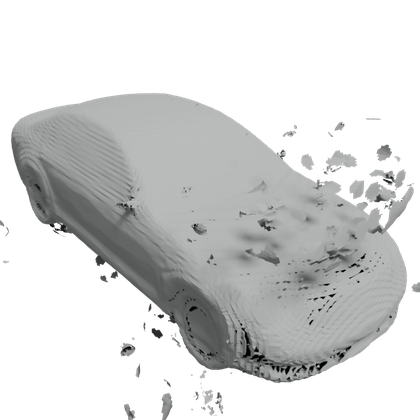}
    \hfill
    \includegraphics[width=0.137\linewidth]{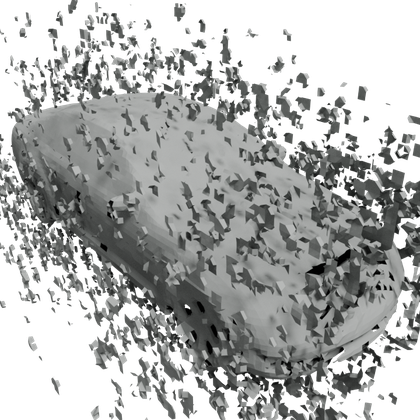}
    \hfill
    \includegraphics[width=0.137\linewidth]{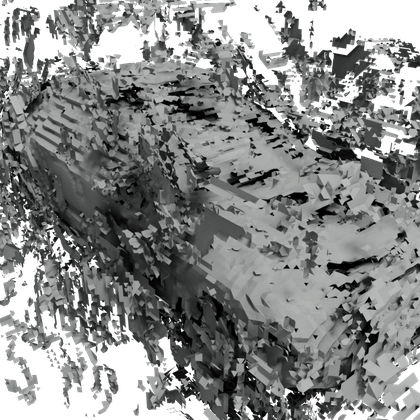}
    \hfill
    \includegraphics[width=0.137\linewidth]{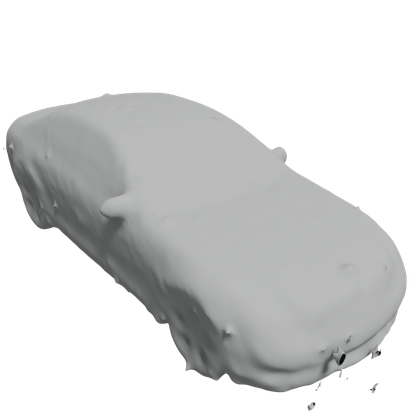}
    \hfill
    \includegraphics[width=0.137\linewidth]{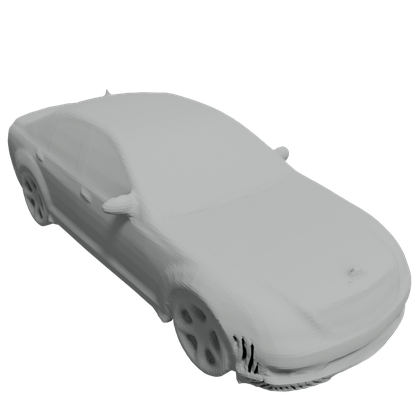}
    \\

    \makebox[0.137\linewidth]{\small(a) \small{Input}}
    \hfill
    \makebox[0.137\linewidth]{\scriptsize (b) \scriptsize {CAP-UDF}}
    \hfill
    \makebox[0.137\linewidth]{\scriptsize (c) \scriptsize{LevelSetUDF}}
    \hfill
    \makebox[0.137\linewidth]{\small(d) \small{GeoUDF}}
    \hfill
    \makebox[0.137\linewidth]{\small(e) \small{DUDF}}
    \hfill
    \makebox[0.137\linewidth]{\small(f) \small Ours}
    \hfill
    \makebox[0.137\linewidth]{\small(g) \small{GT}}
    \caption{More visual  results on the synthetic datasets with outliers.}
    \label{fig:more-outlier}
\end{figure*}

As shown in \cref{fig:more-clean} and \cref{fig:more-outlier}, we provide more visual comparisons on the DeepFashion3D and ShapeNetCars dataset, using point clouds containing noise and outliers.
Our framework is able of reconstructing reasonable surfaces even with 50\% outliers, as shown in Fig.~\ref{fig:more-outliers}. 
Furthermore, our approach can reconstruct high-quality geometry from point clouds containing both noise and outliers.
We also conduct experiments on large scanned scenes to evaluate our algorithm as shown in \cref{fig:scene-reconstrustion}.

\begin{figure*}[h]
    \centering

    \label{fig:more-noise}
\end{figure*}

